%% file: main.tex
\documentclass[runningheads]{llncs}

\usepackage{eccv}

\usepackage{graphicx}
\usepackage{booktabs}

\usepackage{times}

\usepackage[pagebackref,breaklinks,colorlinks,citecolor=eccvblue]{hyperref}

\usepackage[utf8]{inputenc} 
\usepackage[T1]{fontenc}    
\usepackage{url}            
\usepackage{booktabs}       
\usepackage{amsfonts}       
\usepackage{nicefrac}       
\usepackage{microtype}      
\usepackage{xcolor}         
\usepackage{bbding}         
\usepackage{tabularx}       
\usepackage{comment}
\usepackage{color}

\usepackage{amsmath}
\usepackage{amssymb}
\usepackage{float}
\usepackage[symbol]{footmisc}
\usepackage{multirow}

\usepackage[capitalize]{cleveref}
\crefname{section}{Sec.}{Secs.}
\Crefname{section}{Section}{Sections}
\Crefname{table}{Table}{Tables}
\crefname{table}{Tab.}{Tabs.}

\usepackage{eccvabbrv}

\usepackage{graphicx}
\usepackage{booktabs}
\usepackage{multirow}
\usepackage[accsupp]{axessibility}  

\usepackage{orcidlink}

\begin{document}

\title{MotionChain: Conversational Motion Controllers via Multimodal Prompts} 

\titlerunning{MotionChain}




\author{\textbf{Biao Jiang\inst{1,2}\thanks{Work done while Biao Jiang was a Research Intern with Tencent.}\quad
Xin Chen\inst{2}\thanks{Project lead.} \quad Chi Zhang\inst{2} \quad Fukun Yin\inst{1} \quad Zhuoyuan Li\inst{1} \quad \\ Gang Yu\inst{2} \quad Jiayuan Fan\inst{1}\thanks{Corresponding author.}}
}

\institute{$^{1}$Fudan University \quad $^{2}$Tencent}

\authorrunning{Biao Jiang, Xin Chen et al.}

\maketitle

\begin{center}
    \vspace{-10pt}
    \tt \small \textbf{\href{https://github.com/OpenMotionLab/MotionChain}{https://github.com/OpenMotionLab/MotionChain}}
     \vspace{-10pt}
\end{center}

\newcommand{\fix}{\marginpar{FIX}}
\newcommand{\new}{\marginpar{NEW}}

\begin{abstract}
  \input{sections/abstract}
  
\end{abstract}

\input{sections/intro.tex}

\input{sections/related.tex}
\input{sections/method.tex}
\input{sections/experiment.tex}
\input{sections/discussion.tex}


%
%

\bibliographystyle{splncs04}
\bibliography{egbib}

\newpage
\section*{\centering {\LARGE Appendix}}

\input{sections/appendix.tex}

\newpage

\end{document}

%% file: sections/abstract.tex

Recent advancements in language models have demonstrated their adeptness in conducting multi-turn dialogues and retaining conversational context. 
However, this proficiency remains largely unexplored in other multimodal generative models, particularly in human motion models. 
By integrating multi-turn conversations in controlling continuous virtual human movements, generative human motion models can achieve an intuitive and step-by-step process of human task execution for humanoid robotics, game agents, or other embodied systems.
In this work, we present MotionChain, a conversational human motion controller to generate continuous and long-term human motion through multimodal prompts.
Specifically, MotionChain consists of multi-modal tokenizers that transform various data types such as text, image, and motion, into discrete tokens, coupled with a Vision-Motion-aware Language model.
By leveraging large-scale language, vision-language, and vision-motion data to assist motion-related generation tasks, MotionChain thus comprehends each instruction in multi-turn conversation and generates human motions followed by these prompts.
Extensive experiments validate the efficacy of MotionChain, demonstrating state-of-the-art performance in conversational motion generation, as well as more intuitive manners of controlling and interacting with virtual humans.

%% file: sections/intro.tex
\section{Introduction}
\label{sec:intro}

The success of large language models (LLMs)~\cite{ouyang2022instructgpt,openai2023gpt4,workshop2022bloom,touvron2023llama,touvron2023llama2,taori2023alpaca,zheng2023vicuna} has sparked significant interest in the development of multi-modal language models. These models aim to transfer instruction-following and zero-shot abilities to other modalities tasks, such as image-language models~\cite{liu2023llava,ye2023mplug,zhu2023minigpt,alayrac2022flamingo}, video-language models~\cite{alayrac2022flamingo,li2022blip,li2023blip2,zhang2023videollama}, and 3D-language models~\cite{hong20233dllm,guo2023pointllm}. However, a comprehensive model that can perceive visual input and generate continuous motion through multi-turn conversations has not yet been developed.
Such a multi-modal model would have wide-ranging applications in fields like humanoid robotics, virtual assistants, game agents and so on.

Previous research on human motion has explored various tasks, including motion generation~\cite{petrovich21actor, Guo_2022_CVPR_humanml3d,mdm2022human,chen2023mld,zhang2023generating,jiang2023motiongpt}, motion captioning~\cite{goutsu2021seqgan,chuan2022tm2t,jiang2023motiongpt}, motion prediction~\cite{yuan2020dlow,zhang2021we,ma2022multi,jiang2023motiongpt}, and motion composition~\cite{TEACH:3DV:2022}.
Recent works in text-to-motion~\cite{mdm2022human, zhang2022motiondiffuse,petrovich22temos,chen2023mld} have involved pre-trained language models~\cite{devlin2018bert,radford2021clip} for motion generation. For instance, TEMOS~\cite{petrovich22temos} employs BERT~\cite{devlin2018bert} text embeddings in an end-to-end transformer architecture, while MDM~\cite{mdm2022human} and MLD~\cite{chen2023mld} both utilize text embeddings from CLIP~\cite{radford2021clip} during the conditional diffusion process. On the other hand, MotionCLIP~\cite{tevet2022motionclip} and TMR~\cite{petrovich2023tmr} focus on modeling the coupled relationship between motion and text description, and MotionGPT~\cite{jiang2023motiongpt} introduces a motion-language model that represents human motion and language in one unified vocabulary.
However, these above methods treat all tasks as a one-turn conditioned generation, lacking contextual understanding and multi-turn continuous generation abilities.
Therefore, we construct a Vision-Motion language model, integrating multi-turn conversations and continuous human motions.

\begin{figure*}[t]
\centering
\includegraphics[width=1\textwidth]{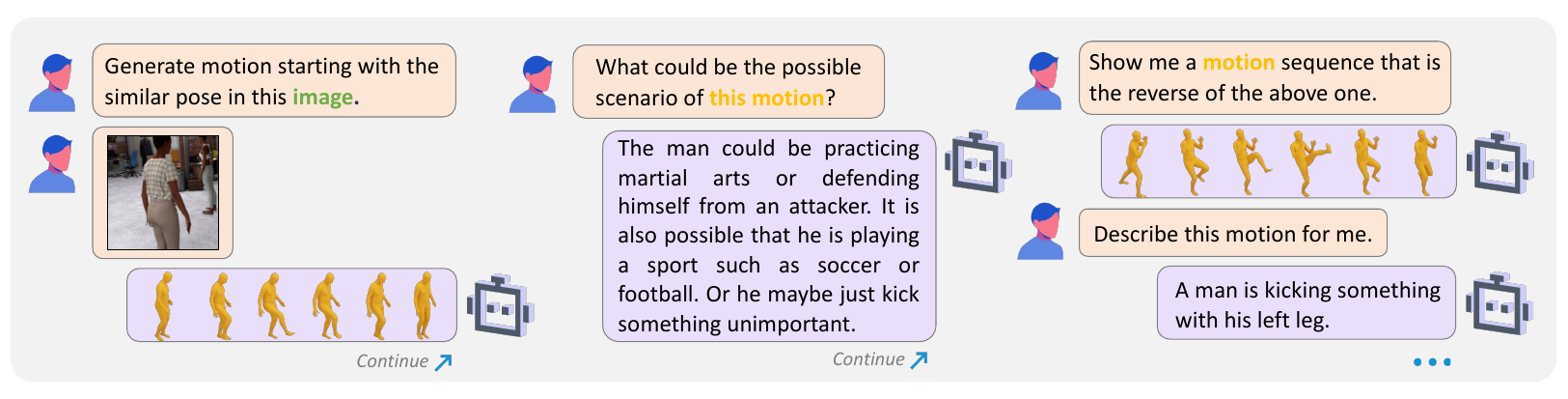}
\caption{MotionChain can interpret instructions from multi-turn conversations and generate human motions or textual answers based on text, motion, or image inputs. We provide the conversation results in image-conditioned motion generation (1st column), motion reasoning (second column), motion editing (third column), and motion translation (third column), with each subsequent turn informed by all previous conversations. Left-to-right represents the temporal order. 
}
\vspace{-10pt}
\label{fig:teaser}
\end{figure*}

Two crucial challenges need to be addressed in this conversational motion generation. The first challenge is to contextually generate human motion in a continuous manner, resembling the way real humans move. The second challenge is the scarcity of text-motion paired datasets compared to datasets with pairs of image-language ~\cite{changpinyo2021cc,schuhmann2022laion}, image-pose~\cite{ordonez2011im2text,kocabas2021spec,mono-3dhp2017,patel2021agora} and video-motion~\cite{ionescu2013human36m,varol17,cai2021gtahuman,bazavan2021hspace,black2023bedlam}.
Fortunately, both human motion and language are sequential and can be continuously "written".
Building upon this observation, we employ the general vision-language instruction-tuning approach 
to enable conversational motion generation and question-answering through multi-modal instructions.
By integrating image, motion, and language data and encoding them into tokens, the relationship between these three modalities becomes more evident.
Therefore, with the advent of large vision-motion and vision-language data, Vision-Motion-language pre-training can enhance the performance of motion-related tasks.

In this study, we introduce MotionChain, a comprehensive framework that integrates vision, motion, and language. MotionChain leverages large-scale vision-language data, video-motion data, and the strong language generation abilities of pre-trained language models to assist in motion-related generation tasks.
To enable MotionChain to comprehend and generate human-like motions, we first train a motion-specific vector quantized variational autoencoder (VQ-VAE) model. This model constructs a "motion vocabulary" similar to the English word vocabulary and converts raw motion data into a sequence of motion tokens.
To incorporate vision inputs into MotionChain, we then introduce a specialized vision tokenizer that connects a pre-trained vision encoder to the language model. This tokenizer converts image data into visual tokens within the language-motion "words" embedding space. These tokens are then processed by a pre-trained language model~\cite{raffel2020t5,chung2022flant5,touvron2023llama,touvron2023llama2}, which learns the relationship between image, motion and language.
To enable conversational generation, we construct a multi-modal motion conversation dataset based on the existing text-motion dataset~\cite{Guo_2022_CVPR_humanml3d} and video-motion dataset~\cite{black2023bedlam}. We then train the language model using our multi-modal conversation dataset to learn the correlation between the three modalities. Extensive experiments demonstrate that MotionChain achieves state-of-the-art performance in multiple motion-related tasks.

We summarize our contributions as follows: (1) We propose MotionChain, a unified vision-motion-language generative pre-trained model, which performs conversational generation tasks via multi-modal inputs with language models.
(2) We introduce a motion composition technique, to generate 3D human motions following the temporal order of instructions.
(3) We propose a multi-modal motion conversation benchmark, wherein MotionChain achieves competitive performance across diverse motion tasks

%% file: sections/related.tex
\section{Related Work}
\label{sec:related}

\textbf{Human Motion Modeling.} 
There have been numerous attempts to model the relationship between 3D human motion and multiple modalities including incomplete motion ~\cite{yuan2020dlow, zhang2021we, ma2022multi,mdm2022human,jiang2023motiongpt},  action~\cite{petrovich21actor, guo2020action2motion,jiang2023motiongpt,chen2023mld,mdm2022human,TEACH:3DV:2022}, text~\cite{Guo_2022_CVPR_humanml3d,chuan2022tm2t,zhang2022motiondiffuse,mdm2022human,ahuja2019language2pose,kim2022flame,jiang2023motiongpt,lu2023humantomato,petrovich2023tmr,petrovich22temos,zhang2023remodiffuse,guo2023momask}, image~\cite{HMR18,guler2019holopose,zhang2019danet,zhang2023pymafx,goel20234dhumans} and video~\cite{VIBE_CVPR2020,rajasegaran2022tracking,goel20234dhumans,black2023bedlam,choudhury2023tempo}. 
Text-to-motion is one of the most important motion generation tasks, due to the user-friendly and convenient language input. MDM~\cite{mdm2022human}, MotionDiffuse~\cite{zhang2022motiondiffuse} and MLD~\cite{chen2023mld} proposes a diffusion-based generative model~\cite{ho2020denoising,song2020denoising,stable_diffusion} to generate motions conditioned on different inputs.
TM2T~\cite{chuan2022tm2t} and T2M-GPT~\cite{zhang2023generating} investigate a generative framework based on VQ-VAE~\cite{van2017vqvae,razavi2019vqvae2} and generative transformer for motion generation. 
Motion completion task generates motion conditioning on partial motions, such as classical motion prediction \cite{yuan2020dlow, zhang2021we, ma2022multi} or motion in-between \cite{mdm2022human}, which generates the intermediate motion while the first and last parts are fixed. 
TEACH~\cite{TEACH:3DV:2022} proposes a past-conditioned transformer model that generate motion from sequence of actions autoregressively.
Apart from motion generation, there is also work investigating other modalities of generation from motion. 
Two statistical models~\cite{takano2015statistical} and recurrent networks~\cite{yamada2018rae,plappert2018learning} are learned in mapping motions to language.
TM2T~\cite{chuan2022tm2t} proposed a new motion representation that compresses motions into a short sequence of discrete variables, then uses a neural translation network to build mappings between two modalities. 
In contrast to the above methods limited to only several tasks, MotionGPT~\cite{jiang2023motiongpt} treats human motion as a foreign language and leverages language understanding and zero-shot transfer abilities of pre-trained language models.

\textbf{Character Control and Animation.}
Character control involves generating interactive motion sequences based on user instruction signals. One kind of approach~\cite{rose1998verbs,kovar2023motiongraphs,min2012motiongraphs++} is to construct a graph representing transitions between motion clips and plan motion using graph search.
Considering the limitations of these graph-based approaches in coarse discreteness, alternative methods like frame blending and concatenation~\cite{lee2010motion}, low-dimensional latent space learning~\cite{levine2012continuous}, motion matching~\cite{clavet2016motion} proposed for embedding the task in the feature and ~\cite{starke2021neural} do the similar thing through hierarchical setup.
Although the control signals for motion control and character animation are different from the instructions in text-to-motion tasks, we still recognize textual commands of conventional human motion generation as a boost for intuitive character control.

\begin{figure*}[t]
\centering
\includegraphics[width=\linewidth]{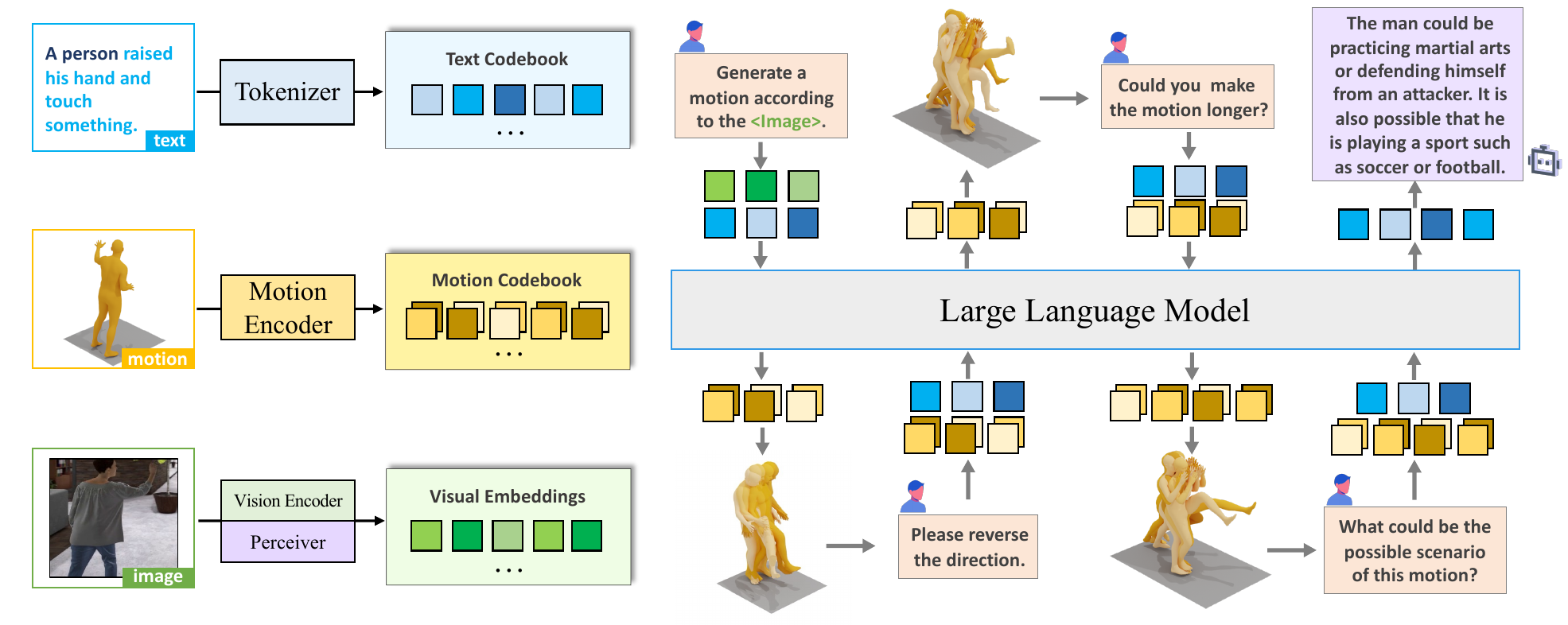}
\caption{Method overview: MotionChain consists of a motion tokenizer $\mathcal{V_M}$ (~\cref{sec:method:tokenizer}), a vision tokenize $\mathcal{V_I}$ (r~\cref{sec:method:tokenizer}) and a vision-motion-aware language model (\cref{sec:method:lm}). By leveraging motion tokens generated by $\mathcal{V_M}$, alongside visual language token embeddings projected by vision tokenizer $\mathcal{V_I}$, and text tokens by text tokenizer, MotionChian achieves a unified learning paradigm for both motion and linguistic data.}
\vspace{-15pt}
\label{fig:pipeline}
\end{figure*}

%
\textbf{Multi-Modal Language Models.}

In the field of computer vision, there has been a recent surge of interest in multi-modal models that can process text along with other modalities, including images, audio, and videos~\cite{li2022blip,huang2023language,girdhar2023imagebind,xu2021videoclip}. CLIP~\cite{radford2021clip} is an example of such a model, which learns a semantic latent representation that connects images with corresponding language descriptions.
While language models have achieved success in various tasks, the development of multi-modal language models capable of handling human motion is still limited. Existing works in computer vision can be broadly categorized into two classes. The first consists of end-to-end trained models explored separately for specific research topics. For example, tasks like vision-language navigation~\cite{anderson2018vision,brooks2023instructpix2pix} and Habitat~\cite{szot2021habitat} require embodied AI agents to follow natural language instructions and take actions to accomplish goals in visual environments. InstructPix2Pix~\cite{brooks2023instructpix2pix} in image translation enables agents to edit images based on human instructions.
The second involves systems that coordinate various models using approaches like LangChain or LLMs~\cite{ouyang2022instructgpt}. Examples of such systems include Visual ChatGPT~\cite{wu2023visualchatgpt}, X-GPT~\cite{zou2023xgpt}, and MMREACT~\cite{yu2022mmreact}. While these methods focus on building instruction-following agents, we aim to develop an end-to-end trained multimodal model that can perform conversational motion generation tasks via multi-modal inputs with language models.

%% file: sections/method.tex
\section{Methods}
\label{sec:method}

\begin{figure*}[t]
\centering
\includegraphics[width=0.8\linewidth]{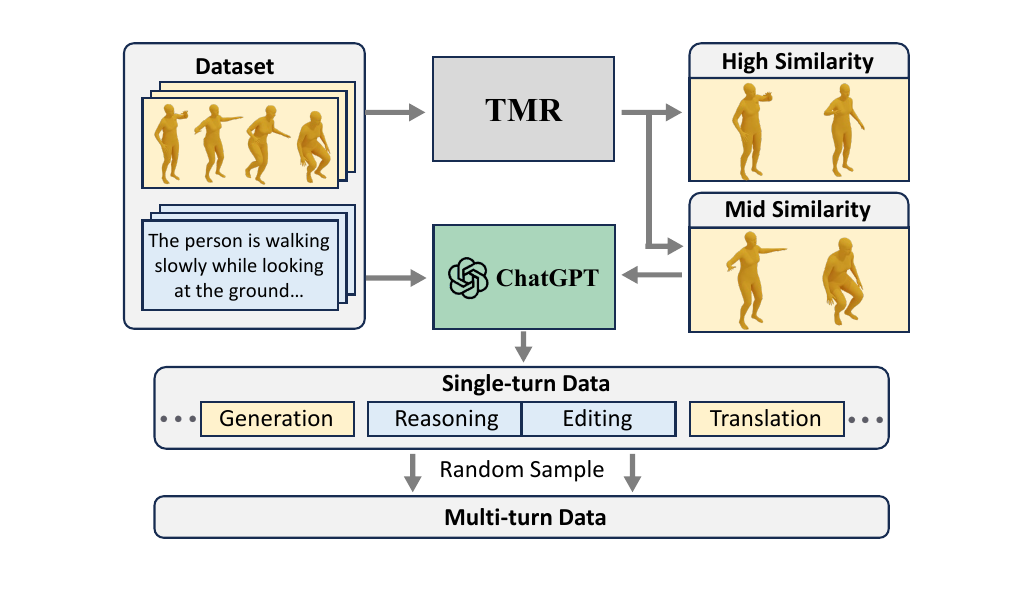}
\vspace{-5pt}
\caption{Data collection overview: Our initial step in collecting the motion reasoning data involves the utilization of human motion captions derived from an existing text-motion dataset. Subsequent to this, the text-motion retrieval model TMR~\cite{petrovich2023tmr} aids in the segmentation of motion pairs into categories based on the similarity between them. With the assistance of ChatGPT, we proceed to craft motion editing task data that correspond to these categorized similarity levels. Incorporating both motion reasoning and editing single-turn tasks, as well as the extensive 14 tasks delineated in \cite{jiang2023motiongpt}, we construct a rich multi-modal multi-turn conversation dataset. }
\vspace{-25pt}
\label{fig:data}
\end{figure*}

To leverage large language data, vision-language data, and vision-motion data for assisting motion-related tasks, we propose a motion-language-vision framework called MotionChain. The framework, as depicted in \cref{fig:pipeline}, consists of \textbf{a multi-modal tokenizer} that converts various types of data (text, image, and motion) into discrete tokens (\cref{sec:method:tokenizer}), and a \textbf{vision-motion-aware Language model} that comprehends information from different modalities and generates corresponding answers based on input instructions (\cref{sec:method:lm}). Additionally, to simultaneously understand data from multiple modalities, we employ \textbf{a multi-stage training strategy} (\cref{sec:method:strategy}) for the training of the multi-modal tokenizer and the motion-language-vision framework.

We first introduced the multi-modal tokenizer, which comprises three branches for processing textual, image, and motion inputs. For textual inputs $w^{1:N} = \{w^i\}$ of length $N$ that describes a motion-related question or demand, we employ the SentencePiece model \cite{kudo2018sentencepiece} used in previous works \cite{raffel2020t5, chung2022flant5, touvron2023llama, touvron2023llama2}, which has a vocabulary size of $K_t$ and is trained on a large number of language datasets. The motion branch consists of a motion encoder $\mathcal{E_M}$ that encodes a motion sequence $m^{1:M} = \{x^i\}$ of $M$ frames into $L$ motion tokens $z^{1:L} = \{z^i\}$, where $L = M/l$ and $l$ represents the temporal downsampling rate on motion frames. It also includes a motion decoder $\mathcal{D_M}$ that can decode motion tokens back to human motion $\hat{m}^{1:M}$. The vision branch processes the input image $X$ with a pre-trained CLIP visual encoder and a learnable linear projection that follows it, converted into language token embeddings $H_q$. 
Given a textual sentence $w^{1:N}$, a sequence of motion $m^{1:M}$, and an image condition $X$, all encoded as language tokens, our vision-motion-aware language model is designed to produce an answer comprising $L$ tokens, denoted as $\hat{x}^{1:L} = \{\hat{x}^i\}$. These output tokens can represent either motion sequences $\hat{x}_m^{1:L}$ or textual descriptions $\hat{x}_t^{1:L}$, which integrate both human motion $\hat{m}^{1:M}$, and text $\hat{w}^{1:L}$  within the given context.

\subsection{Data Collection}
\label{sec:method:data}
With the emergence of text-conditioned motion generation tasks, datasets like KIT~\cite{Plappert2016kit}, BABEL~\cite{BABEL:CVPR:2021}, HumanML3D~\cite{Guo_2022_CVPR_humanml3d} and the more recent Motion-X~\cite{lin2023motionx} have been developed. However, these datasets predominantly offer text labels as simple action phrases or captions. Building upon these foundations,  MotionGPT~\cite{jiang2023motiongpt} introduces an instruction-based motion-language dataset that encapsulates 14 core tasks, including motion prediction, translation, and editing, through thousands of instruction templates in a unified format. Despite this advancement, MotionGPT's data lack a deep engagement with the nuances of human motion analysis and are limited to single-turn generation tasks without incorporating contextual memory.
Inspired by the recent success of GPT models across text-annotation tasks~\cite{gilardi2023gptlabel}, image-annotation tasks~\cite{liu2023llava}, 3D-annotation tasks~\cite{hong20233dllm}, we propose a data collection methodology integrates the capabilities of existing LLMs like ChatGPT~\cite{ouyang2022instructgpt}, with the text-motion retrieval model TMR~\cite{petrovich2023tmr} to facilitate motion conversation data collection. In addition to the 14 motion-related tasks in MotionGPT~\cite{jiang2023motiongpt}, we introduce tasks centered around motion reasoning and motion editing, leveraging contextual insights for a deeper motion analysis. 

Utilizing ChatGPT~\cite{ouyang2022instructgpt}, we initiate the collection of motion reasoning data using human motion captions from the text-motion dataset~\cite{Guo_2022_CVPR_humanml3d}, starting with manually designed example queries that explore the contextual scenarios surrounding motions, possible preceding or succeeding actions, the subjects' roles, and the tools or equipment involved, etc.
Following this, we employ TMR~\cite{petrovich2023tmr} for categorizing motions from the dataset into varying similarity levels. For medium-similarity motion pairs, we utilize ChatGPT~\cite{ouyang2022instructgpt} to generate motion editing directives that enable the transformation of one motion to another. For motions of high similarity, we manually devise tasks aimed at editing their lengths, further enriching the dataset's versatility and analytical scope.

After the collection of single-turn generation tasks, we progress to develop multi-turn conversation data. This involves the deliberate association of initial motion generation tasks with a variety of follow-up tasks randomly chosen among motion translation, reasoning, editing, etc. Following~\cite{zheng2023vicuna}, we construct our conversation data in a structured format, as depicted below:

\begin{quote}

\small$X_{\text{system-message}}$ 

USER: $X_v$ $X_s^1$ ASSISTANT: \textcolor{Green}{$X_a^1$ \small{$<$/s$>$}}

USER: $X_v$ $X_s^2$ ASSISTANT: \textcolor{Green}{$X_a^2$ \small{$<$/s$>$}}

USER: $X_v$ $X_s^3$ ASSISTANT: \textcolor{Green}{$X_a^3$ \small{$<$/s$>$}} ...

\end{quote}
Where $X_v$ is defined as the vision language token embeddings, processed via the visual tokenizer. $X_s^i$ and $X_a^i$ are used to denote the source inputs and target answers for each round $i$, respectively. Both sets of tokens originate from the integrated motion-language vocabulary $V$, which includes motions, texts, or a blend thereof. The dataset exhibits variability in the number of generation turns up to 10; for the sake of clarity, we present only three examples herein. MotionChain is trained to predict answers, incorporating a learning mechanism that determines whether to stop generation by outputting end of sentence flag {$<$/s$>$} based on the current instruction and all preceding questions and answers. In the computation of the loss, as defined in \cref{eq:loss:lm}, only the \textcolor{Green}{green tokens} are utilized.

\subsection{Multi-modal Tokenizer}
\label{sec:method:tokenizer}
\textbf{Motion tokenizer}, denoted as $\mathcal{V_M}$, is based on the architecture of Vector Quantized Variational Autoencoders (VQ-VAE) utilized in previous studies \cite{van2017vqvae, siyao2022bailando, chuan2022tm2t, zhang2023generating,yao2023moconvq,jiang2023motiongpt,wang2023t2mhifi,guo2023momask}. Once pre-trained, it can represent motion using discrete tokens, facilitating the integration of motion and language. The Motion tokenizer consists of a motion encoder $\mathcal{E_M}$ and a motion decoder $\mathcal{D_M}$. Initially, the motion encoder $\mathcal{E}$ applies 1D convolutions to the motion features $m^{1:M}$ along the temporal dimension to obtain latent vectors $\hat{z}^{1:L} = \mathcal{E_M}(m^{1:M})$. Subsequently, the latent vectors $\hat{z}$ are quantized and transformed into a collection of codebook entries $z$. The learnable codebook $Z = \{{z}^i\}_{i=1}^{K} \subset \mathbb{R}^{d}$ comprises $K$ latent embedding vectors, each with a dimension of $d$. The quantization process $Q(\cdot)$ replaces each row vector $b$ with its ne,arest codebook entry $b_k$ in $Z$, which can be expressed as:

\begin{equation}
z_i = Q(\hat{z}^i) := {\arg \min }_{z_k \in Z}\left\|\hat{z_i} - z_k\right\|_2.
\end{equation}

We assign ${s}^i$ as the index number of motion tokens ${z}^{1:L}$, so motion tokens $z^{1:L}$ can be represented as a sequence of indices ${s}^{1:L}=\{{s}^i\}_{i=1}^{L}$. The motion decoder $\mathcal{D_M}$ can project ${z}^{1:L}=\{{z}^i\}_{i=1}^{L}$ back to the raw motion space, resulting in the motion $\hat{m}^{1:M}$ with $M$ frames.
Following~\cite{chuan2022tm2t, zhang2023generating,jiang2023motiongpt,wang2023t2mhifi,guo2023momask}, we adopt three distinct loss functions when training the motion tokenizer: 
\begin{equation}
    \mathcal{L}_\mathcal{V} = \mathcal{L}_{r} + \mathcal{L}_{e} + \mathcal{L}_{c}\label{eq:loss:vq}
\end{equation}
where $\mathcal{L}_{r}$ denotes reconstruction loss, $\mathcal{L}_{e}$ denotes the embedding loss, and $\mathcal{L}_{c}$ denotes commitment loss. 

During multi-turn motion generation, the motion continuity between turns is achieved through our motion decoder, which links the motion of the current turn with that of the preceding ones. Taking the composition of two motions as an example: we concatenate the past motion tokens, denoted as ${z}_p^{1:L_p}$, with the tokens representing the current motion, ${z}_c^{1:L_c}$. This concatenated sequence of tokens is subsequently decoded into a comprehensive set of continuous motion features, represented as $m_{\text{whole}}^{1:M_{\text{whole}}}$, as depicted below:
\begin{equation}
    {z}_{\text{whole}}^{1:(L_p+L_c)} = [{z}_p^{1:L_p},{z}_c^{1:L_c}]. 
\end{equation} 
Similarly, this framework is adept at executing composition tasks involving an array of motions. The comparison results in \cref{tab:able: composition} demonstrate that our motion tokenizer could effectively perform motion composition tasks.

\textbf{Visual Tokenizer} accepts both image ($X_I$) and video($X_V$) as inputs. For the image input, we employ the CLIP visual encoder that is pre-trained on image-text pairs to derive visual feature $Z_I$. These features are then projected into language token embeddings $X_v$ via a linear layer like previous wrok~\cite{liu2023llava}.
For videos, each sampled frame is encoded through the CLIP visual encoder resulting in a 3D spatiotemporal feature matrix with additional temporal embeddings. Inspired by  ~\cite{alayrac2022flamingo}, we introduce a perceiver module. This module integrates a transformer equipped with fixed-length, learnable queries, that dynamically engage with these visual features to synthesize consistent output. Subsequently, a linear layer, applied similarly to image inputs, uses a trainable matrix $W$ to map $Z_I$ to visual token embeddings $X_v$, maintaining consistency in the dimensionality with the language model's word embedding space.

\begin{figure}[t]
  \centering
  \begin{subfigure}[t]{0.325\textwidth}
    \centering
    \includegraphics[width=0.9\linewidth]{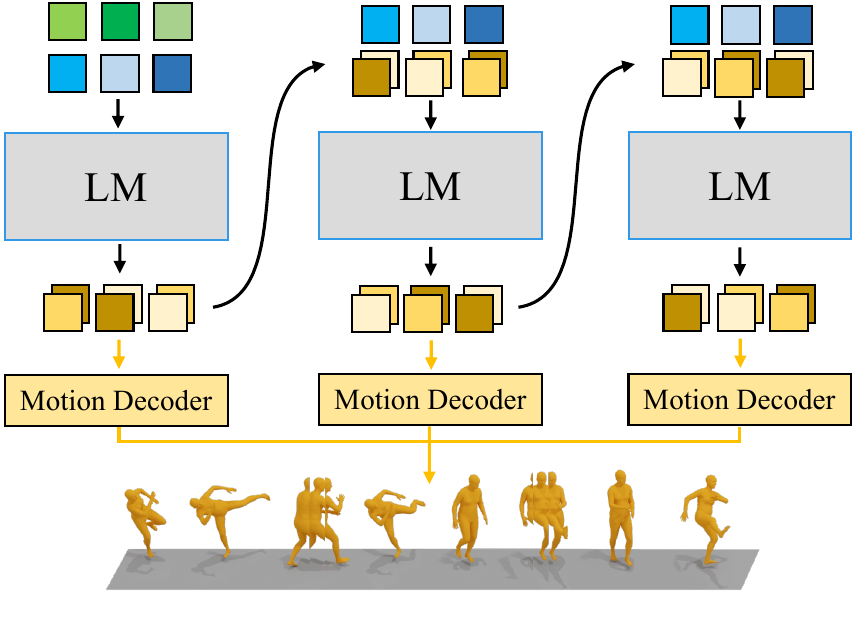}
    \caption{Independent}
    \label{fig:vq-a}
  \end{subfigure}
  \begin{subfigure}[t]{0.325\textwidth}
    \centering
    \includegraphics[width=0.9\linewidth]{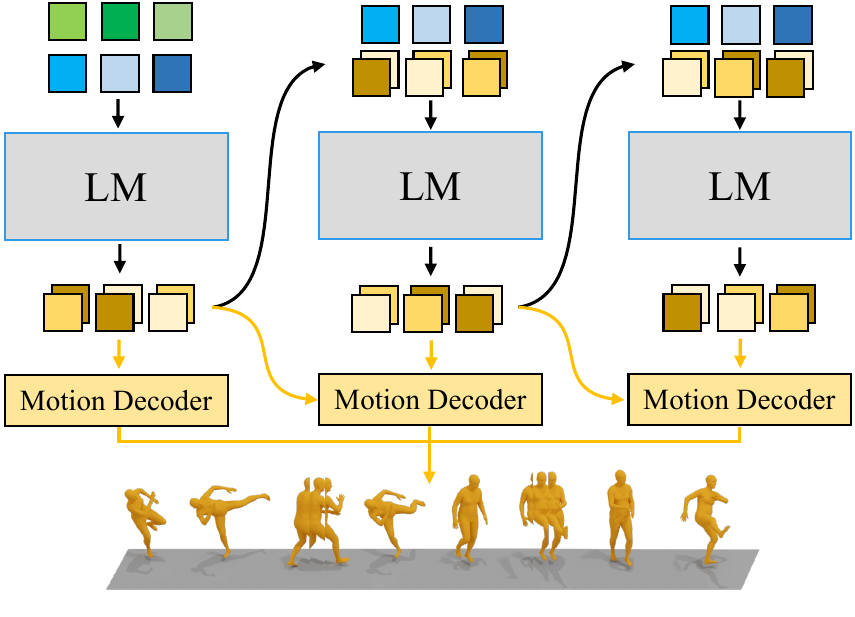}
    \caption{Past-condition}
    \label{fig:vq-b}
  \end{subfigure}
  \begin{subfigure}[t]{0.325\textwidth}
    \centering
    \includegraphics[width=0.9\linewidth]{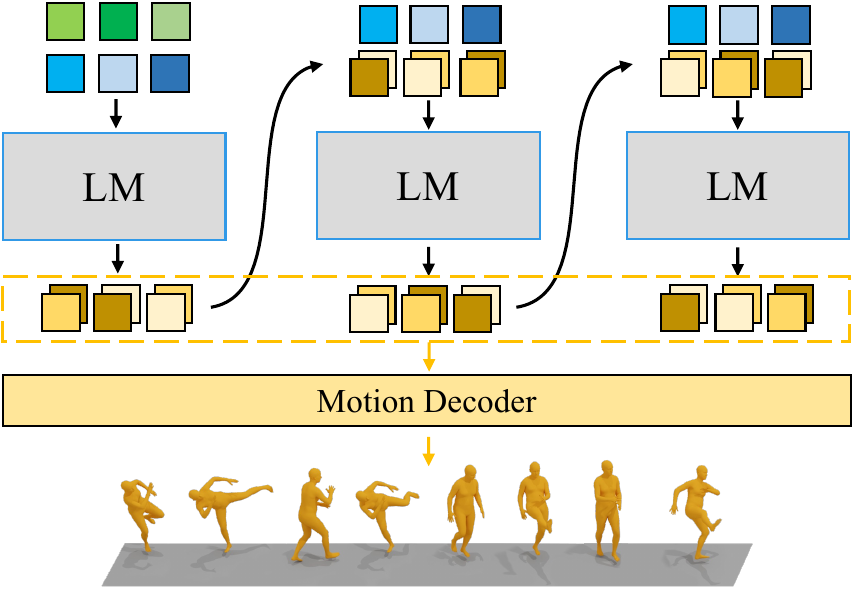}
    \caption{Tokens-joint}
    \label{fig:vq-c}
  \end{subfigure}
  \caption{Motion Composition Variants: We illustrate the baselines for motion composition during multi-turn motion generation (a). independent decoding each turn (b). separate decoding conditioned on the last few tokens from the prior turn (c). decoding with joint motion tokens. Green tokens stand for image condition, blue tokens stand for textual instruction, and orange tokens stand for human motions.}
  \label{fig:composition}
  \vspace{-15pt}
\end{figure}

\subsection{Motion-aware Language Model}
\label{sec:method:lm}
Language models such as Llama~\cite{touvron2023llama,touvron2023llama2} and T5~\cite{raffel2020t5,chung2022flant5} employ the SentencePiece~\cite{kudo2018sentencepiece} model to encode textual inputs into WordPiece tokens, utilizing a $K_t$ word piece vocabulary. Unlike prior text-to-motion~\cite{chuan2022tm2t, chen2023mld, zhang2023generating, zhang2023remodiffuse} and motion-to-text~\cite{chuan2022tm2t} methods that process text and motion separately, we merge the text vocabulary $V_t=\{{v}_t^i\}_{i=1}^{K_t}$ with the motion vocabulary $V_m=\{{v}_m^i\}_{i=1}^{K_m}$, maintaining the motion tokenizer's codebook $Z$ order and including special tokens for boundary demarcation. This creates a unified vocabulary $V = \{V_t, V_m\}$, enabling the formulation of motion-centric tasks in a universal template, where inputs and outputs share the same vocabulary. For visual input, our visual tokenizer converts images or videos into visual token embeddings $X_v$, aligning with the language model ~\cite{kudo2018sentencepiece, raffel2020t5, ouyang2022instructgpt, touvron2023llama} token space for integrated representation.


For single conditioned generation tasks, our input comprises a sequence of $N$ length tokens ${X_s}=\{{x_s}^i\}_{i=1}^{N}$, where $x_s\in \{V_t, V_m\}$ representing either text, motion, or a combination thereof, drawn from the unified vocabularies. In cases involving image inputs, visual tokens $X_v$ are interspersed at the beginning of the source tokens sequence, forming $[X_v, X_s]$. Subsequent interaction rounds generate target answer tokens $X_a$.
To facilitate iterative result generation and content retention, our framework generates multi-turn conversation data $(X_v, X_s^1, X_a^1, X_s^2, X_a^2,\cdots, X_s^T, X_a^T)$, with $T$ indicating the total turn count. Notably, visual tokens are consistently placed at the forefront of the initial turn's source tokens. 
The processing sequence is organized such that to predict target answer tokens autoregressively, as shown in \cref{fig:pipeline}. Source tokens are processed by the transformer to predict the next token's probability distribution, formulated as:
\begin{equation}
    p_\theta(X_a \mid X_v, X_s )=\prod_i p_\theta\left(x_a^i \mid X_v, X_{s,<i}, X_{a,<i}\right)
    \label{eq:lm:autoregressive}
\end{equation}
with $\theta$ indicating trainable parameters, and $X_{s,<i}, X_{s,<i}$ the sequences of source and preceding target tokens. The training objective is maximizing the log-likelihood of distribution:
\begin{equation}
    \mathcal{L}_{LM}=-\sum_{i=0}^{L_t-1} \log p_\theta\left(x_a^i \mid X_v, X_{s,<i}, X_{a,<i}\right) .
    \label{eq:loss:lm}
\end{equation}
By optimizing this objective, MotionChain captures the complex interrelations among images, motion, and text, facilitating accurate target "word" generation.

During the inference phase, target tokens are recursively sampled from the model's predicted distribution $p_\theta\left(\hat{x_a}^i \mid X_v, X_s, \hat{X}_{a,<i} \right)$, ceasing with the appearance of a special end token. This strategy facilitates a step-by-step target sequence generation, where each token's probability is conditioned on all previous turns' sources and targets and current source input.

\subsection{Training Strategy}
\label{sec:method:strategy}
To facilitate the integration of image and motion comprehension within the language modeling context, we adopt a 3-stage training strategy. (1) The initial stage involves pre-training the motion tokenizer on a corpus of human motion data, in line with ~\cite{jiang2023motiongpt}. This process establishes the motion vocabulary $V_m$, which serves as a foundation for encoding human motions as a series of discrete tokens. (2) Subsequently, the motion tokenizer remains frozen while we connect the visual tokenizer to the language model framework. This integration is supported by a suite of supervised objectives, including text-to-motion, motion-to-text, and image-based motion generation, aiming to learn the intricate relationships between images, motion, and language. (3) The final stage involves instruction tuning, and refines the model's capabilities through the application of prompt-based instructions. These instructions are framed within multi-turn conversation sequences, as detailed in \cref{sec:method:lm}, to expanded range of motion-related tasks.

\textbf{Training of Motion Tokenizer.} 
The initial step involves training the motion tokenizer, guided by the loss objective in Equation \ref{eq:loss:vq}. This stage enables the tokenizer to represent human motion sequences $\hat{x}^{1:L}$ as discrete motion tokens, a key step for merging motion data with textual information seamlessly. Once optimized, the motion tokenizer remains frozen.

\textbf{Motion-language Pre-training Stage.} 
Leveraging recent developments in language modeling \cite{raffel2020t5,chung2022flant5,touvron2023llama,touvron2023llama2,zheng2023vicuna} pre-trained on natural language datasets and then fine-tuned with instruction-based phrasing \cite{chung2022flant5,ouyang2022instructgpt}. To augment the model's ability to discern relationships between images and human motions, we first pre-train our MotionChain using a mix of language, image, and motion datasets. Following the stage 1 training of the motion tokenizer, we have established a unified motion-language vocabulary $V={V_t, V_m}$, capable of representing motions in discrete token form. Moreover, we maintain the visual encoder's weights in the visual tokenizer as fixed, while the linear projection weight $W$ is jointly optimized with the language model. During this stage, the model undertakes three fundamental single-turn modality translation tasks: text-to-motion, motion-to-text, and image-conditioned motion generation, as outlined in \cref{sec:method:data}. The primary objective is to maximize the likelihood of the model according to the loss function specified in \cref{eq:loss:lm}, thereby letting the model understand the relationship between language, vision conditions, and motions.

\textbf{Instruction Tuning Stage.} 
As described in \cref{sec:method:data}, we construct a multi-modal, multi-task, and multi-turn motion conversation dataset by augmenting existing text-to-motion~\cite{Guo_2022_CVPR_humanml3d} and human mesh reconstruction datasets~\cite{black2023bedlam} with targeted instructional prompts and leverage the capabilities of LLMs~\cite{ouyang2022instructgpt} and the text-motion retrieval model~\cite{petrovich2023tmr} for motion reasoning and editing tasks. The efficacy of instruction tuning, as evidenced across language models~\cite{chung2022flant5,ouyang2022instructgpt,zheng2023vicuna,liu2023llava}, is well-established, yielding enhancements in model performance across a wide range of tasks. After instruction tuning, MotionChain can handle more motion-related tasks including the proficient handling of previously unseen tasks

%% file: sections/experiment.tex
\section{Experiments}
\label{sec:experiments}
We evaluate the proposed MotionChain encompasses comprehensive comparisons across both one-turn motion-related tasks and multi-turn motion generation tasks. Firstly, we provide details of the dataset settings, evaluation criteria, and implementation details as specified in \cref{sec:comp:detail}. Subsequently, comparative analyses are presented, focusing on the motion reasoning task (\cref{sec:comp:qa}) and the temporal motion composition task (\cref{sec:comp:t2m2}). In \cref{sec:abl}, we evaluate the choice of motion composition technique and different architectures of vision tokenizer.

\subsection{Experimental Setup}
\label{sec:comp:setup}
\textbf{Datasets.} 
For one-turn motion reasoning tasks, the study employs our proposed multi-modal multi-turn conversation dataset upon HumanML3D~\cite{Guo_2022_CVPR_humanml3d} with 44,970 sequence-level textual descriptions for 14,616 motion sequences obtained from AMASS \cite{AMASS_ICCV2019} and HumanAct12~\cite{guo2020action2motion}. The datasets are divided into training, testing, and validation sets with a ratio of $0.8: 0.15:0.05$.
To evaluate the multi-turn motion generation task, we focus on BABEL~\cite{BABEL:CVPR:2021} that provides textual descriptions for the motions in the AMASS~\cite{AMASS_ICCV2019} with annotated segments that overlap in each sequence, which allows evaluating generation of a sequence of motion or actions. We adopt the processed text labels by~\cite{TEACH:3DV:2022} and motion representation of HumanML3D~\cite{Guo_2022_CVPR_humanml3d} which combines joint velocities, positions, and rotations. Following~\cite{TEACH:3DV:2022} we consider pairs of actions for simplicity but MotionChain applies to a sequence of actions or motion of arbitrary length.
For the image-conditioned motion generation task, we mainly focus on BEDLAM~\cite{black2023bedlam}, a large synthetic dataset of realistic moving 3D humans containing more than 200 subjects and 380K frames video and motion pair.

\textbf{Evaluation Metrics} are summarized as four parts. (1) Motion quality:
We adopt Frechet Inception Distance (FID) as the primary metric, FID quantifies the divergence in feature distributions between generated and actual motion sequences. Utilizing feature extractors from prior studies~\cite{Guo_2022_CVPR_humanml3d,lin2023motionx,petrovich2023tmr}, FID measures the distance of feature distributions between the generated and real motions. Following ~\cite{HMR18,vposer_SMPL-X:2019,chen2023mld,black2023bedlam,jiang2023motiongpt}, we also adopt MPJPE, PA-MPJPE to measure global and local errors in millimeters and ACCL for acceleration errors, to evaluate the quality of the reconstructed motions.
(2) Motion Diversity: Utilizing the Diversity (DIV) metric, we calculate variance across motion features to evaluate generation diversity.
(3) Text matching: 
The precision of text-to-motion matches is quantified by the R Precision metric, based on the feature evaluator~\cite{Guo_2022_CVPR_humanml3d,lin2023motionx,petrovich2023tmr}, and includes an analysis of Top 1/2/3 retrieval accuracy. The Multi-modal Distance (MM Dist) quantifies the semantic gap between motions and texts.
(4) Linguistic quality: We follow \cite{chuan2022tm2t} utilizing linguistic metrics from natural language studies, including BLUE~\cite{papineni2002bleu}, Rouge~\cite{lin2004rouge}, Cider~\cite{vedantam2015cider}, and BertScore~\cite{zhang2019bertscore} to evaluate the quality of generated motion captions.
More detailed benchmark information is provided in the supplementary materials.

\label{sec:comp:detail}
\textbf{Implementation Details.} 
We set the codebook of the motion tokenizer as $K\in\mathbb{R}^{512\times1024}$ for most experiments. The motion encoder, denoted as $\mathcal{E_M}$, integrates a temporal downsampling rate, $l=4$. Our vision tokenizer incorporates a frozen Vision Transformer (ViT-L/14)~\cite{radford2021clip} as visual encoder for most experiments. Additionally, for comprehensive ablation studies, we explored the use of both a frozen vision encoder and a Q-former from BLIP-2~\cite{li2023blip2} as a vision tokenizer.
We mainly utilize Flan-T5-base~\cite{chung2022flant5} as the underlying architecture for our language model.
Moreover, all our models employ the AdamW~\cite{loshchilov2017adamw} optimizer with $[\beta_1,\beta_2]=[0.9,0.99]$ for training. The motion tokenizers are trained to utilize a $10^{-4}$ learning rate employing cosine annealing scheduler and a 256 mini-batch size. Our language models based on Flan-T5-base~\cite{chung2022flant5} have a $10^{-4}$ learning rate with cosine annealing scheduler and 16 mini-batch sizes in both the pre-train stage and the instruction tuning stage. The motion tokenizer undergoes 10000 epochs of training, while the language model undergoes 500 epochs during the pre-train stage and another 50 epochs during the instruction tuning stage. Most models are trained on 8 Tesla V100 GPUs.

\subsection{Comparisons on Motion Reasoning.}
\label{sec:comp:qa}
In \cref{sec:method:data}, we introduce a multi-modal motion conversation dataset, enriched with motion reasoning data facilitated by ChatGPT~\cite{ouyang2022instructgpt}. This task evaluates the model's reasoning capabilities with motion reasoning tasks, where a motion sequence or its corresponding textual descriptions serve as inputs.
Our evaluation compares our MotionChain, which integrates motion perception, against contemporary Large Language Models (LLMs) that possess solely textual processing capabilities. The compared LLMs are assessed using their original pre-trained weight. Results in \cref{tab:tm:comp:reasoning}, illustrate that MotionChain exhibits superior motion reasoning proficiency, benefiting from its integrated motion perception.

\begin{table}[t]
\centering

\resizebox{\columnwidth}{!}{
\begin{tabular}{@{}lcccccccccc@{}}
\toprule
Methods
&Params
&$\text{Length}_{\text{avg}}$
&Bleu@1$\uparrow$& 
Bleu@4$\uparrow$&
Rouge$\uparrow$&
Cider$\uparrow$&
BertScore$\uparrow$

\\\midrule
Flan-t5-base~\cite{chung2022flant5} & 250M & $8.34$ & $4.64$ & $1.78$ & $15.32$ & $15.93$ & $3.45$
\\
Flan-t5-large~\cite{chung2022flant5} & 780M & $11.95$ & $12.18$ & $4.83$ & $22.81$ & $15.02$ & $14.19$
\\
Flan-t5-xl~\cite{chung2022flant5} & 3B & $9.09$ & $8.54$ & $4.01$ & $24.89$ & $15.03$ & $18.34$
\\
Llama-2-7b~\cite{touvron2023llama2} & 7B & $130.84$ & $11.12$ & $3.67$ & $19.14$ & $1.04$ & $6.81$
\\
Vicuna-1.5-7b~\cite{zheng2023vicuna} & 7B & $71.49$ & $19.27$ & $7.39$ & $25.75$ & $5.44$ & $19.05$ 
\\
Vicuna-1.5-13b~\cite{zheng2023vicuna} & 13B & $84.74$ & $17.20$ & $6.53$ & $24.18$ & $7.77$ & $18.00$
\\\midrule
MotionChain & 280M & 22.17 & $\boldsymbol{37.92}$ & \textbf{$\textbf{19.19}$} & \textbf{$\textbf{38.05}$} & \textbf{$\textbf{24.53}$} & \textbf{$\textbf{32.24}$}
\\ \bottomrule
\end{tabular}%
}
\vspace{5pt}
\caption{Comparison of motion reasoning on the test set of our conversation dataset. Our proposed MotionChain is fine-tuned on motion reasoning tasks while other methods' results are generated by their pre-trained weight.
\textit{$Length_\textit{avg}$} represents the average words in generated answers to all questions. We adopt metrics commonly used in natural language processing tasks for evaluation. }
\label{tab:tm:comp:reasoning}
\vspace{-20pt}
\end{table}

\subsection{Comparisons on Temporal Composition.}
\label{sec:comp:t2m2}
The temporal motion composition task involves generating a continuous motion sequence from two actions in a time series. We conducted our experiments following the settings in TEACH~\cite{TEACH:3DV:2022} and used the Amass~\cite{AMASS_ICCV2019} subset BABEL~\cite{BABEL:CVPR:2021} validation set. Additionally, we processed the motion in Amass into the format proposed by HumanML3D~\cite{Guo_2022_CVPR_humanml3d} and trained our MotionChain on the action-to-motion task. To compare with TEACH, we initially used an officially provided pre-trained model to sample motion on the validation set 20 times. Subsequently, we post-processed their motion into the HumanML3D format, represented in SMPL~\cite{SMPL2015}. The performance of our MotionChain is summarized in Table \ref{tab:tm:comp:babel}. As evaluating generative models quantitatively is challenging, we also provide qualitative comparisons in the supplementary materials.

\begin{table}[t]
\centering
\begin{tabular}{@{}lcccccccc@{}}
\toprule
{Methods}&
 {Diversity} &{MPJPE$\downarrow$} & {PA-MPJPE$\downarrow$} &{ACCL$\downarrow$}\\
\midrule
Real &
  $15.74^{\pm.149}$ &
  - &
  - &
  - 
  \\ 
  \midrule
Teach~\cite{TEACH:3DV:2022} &
${27.11}^{\pm.159}$ &
$979.21^{\pm.215}$ &
$933.32^{\pm.254}$ &
$23.02^{\pm.018}$
\\
MotionChain &
$43.25^{\pm.159}$ &
$\textbf{276.05}^{\pm6.72}$ &
$\textbf{53.72}^{\pm.580}$ &
$\textbf{7.11}^{\pm0.100}$
\\ \bottomrule
\end{tabular}%
\vspace{5pt}
\caption{Comparison of temporal motion composition on Babel~\cite{BABEL:CVPR:2021}. We evaluate the state-of-the-art motion temporal composition method Teach~\cite{TEACH:3DV:2022} under the 95 \% confidence interval from 20 times running. ($cf.$ \cref{sec:comp:setup} for notations.)
}
\vspace{-15pt}
\label{tab:tm:comp:babel}
\end{table}

\subsection{Ablation Studies}
\label{sec:abl}
MotionChain enables multi-modal motion conversation using two main techniques. The first technique involves generating a smooth sequence of motions by concatenating motion tokens which are then decoded back to motion by motion decoder $\mathcal{D_M}$. The second technique involves processing multi-modal visual input through a vision tokenizer, which consists of a frozen vision encoder and a trainable linear projection. To evaluate the effectiveness of these two designs, we compare them with other variants. For a more comprehensive analysis, detailed ablation studies can be found in the supplementary materials.

\begin{table}[t]
\centering

\begin{tabular}{@{}lccccccc@{}}
\toprule
{Method}&MPJPE$\downarrow$&PA-MPJPE$\downarrow$&ACCL$\downarrow$ & Diversity
\\\midrule
Independent & $350.79$ & $102.97$ & $11.40$ & $ 6.47$
\\ 
Past-condition & $232.46$ & $46.15$ & $6.18$ & $6.01$
\\
Tokens-joint & $\textbf{108.77}$ & $\textbf{18.85}$ & $\textbf{2.26}$ & ${5.56}$ 
\\ \bottomrule
\end{tabular}%
\vspace{5pt}
\caption{
Evaluation of motion composition methods on HumanML3D~\cite{Guo_2022_CVPR_humanml3d}. Here \textit{Independent}, \textit{Past-condition}, and \textit{Tokens-joint} stand for different motion composition varients during multi-turn motion conversation, as illustrated in \cref{fig:composition}.
}
\label{tab:able: composition}
\vspace{-15pt}
\end{table}
\textbf{Motion Composition Mechanism}
\label{sec:abl:composition}
Apart from the jointly token concatenating mechanism, we also evaluate the performance of temporal motion composition through the other motion temporal composition variants Motion-cat: concatenating the motion in final motion level rather than token level. 
Experimental results in \cref{tab:able: composition} show that jointly concatenating motion tokens achieved remarkable performance compared to the other variants. For further information regarding the implementation of the aforementioned vision tokenizer, please refer to the supplementary materials.

\textbf{Image Tokenizeer Architecture.}\label{sec:abl:vt}
MotionChain connects the frozen vision encoder to the language model through a linear layer. However, previous vision-language~\cite{alayrac2022flamingo,li2023blip2} works also demonstrate the effectiveness of other kinds of visual-aligning modules. Here we consider the other two vision tokenizer variants: (a) inspired by \cite{carion2020end,jaegle2021perceiver,alayrac2022flamingo}, we introduce a perceiver module that incorporates a transformer receiving a predefined number of latent input queries. These queries cross-attend to the visual features, enabling effective information exchange. (b) We directly adopt the pre-trained Q-former from BLIP-2~\cite{li2023blip2} to align visual inputs with the language model. 
We evaluate the different architectures under the single human image as the first frame condition and the last frame condition separately.
Experimental results in \cref{tab:abl:vt} show that a lightweight linear projection is sufficient for comprehending the human pose from visual input. 
Additional details about the implementation of the above vision tokenizer can be found in the supplements.

\begin{table}[t]
\centering
\begin{tabular}{@{}lccccccc@{}}
\toprule

\multirow{2}{*}{Architecture}&\multicolumn{2}{c}{First-frame} & \multicolumn{2}{c}{Last-frame}
\\\cmidrule(lr){2-3}\cmidrule(lr){4-5}&
MPJPE $\downarrow$ & PA-MPJPE $\downarrow$ &  MPJPE $\downarrow$ & PA-MPJPE $\downarrow$ \\
\midrule
Q-former & $195.49$ & $86.56$ & $134.73$ & $57.17$
\\
Perceiver & $185.61 $ & $99.21$ & $134.89$ & $57.58$
\\
Linear & $\textbf{144.37}$ & $\textbf{76.48}$ & $\textbf{133.73}$ & $\textbf{56.73}$
\\
\bottomrule
\end{tabular}%
\vspace{5pt}
\caption{Evaluation of vision tokenizer architecture on Bedlam~\cite{lin2023motionx}. We implement three different architectures, including Q-former, Perceiver, and Linear. We evaluate these results with the metrics in motion reconstruction. Additional information regarding the implementation is in the supplementary materials. ($cf.$~\cref{tab:tm:comp:babel} for notations.)}
\label{tab:abl:vt}
\vspace{-15pt}
\end{table}

%% file: sections/discussion.tex
\section{Conclusion and Limitation}
\label{sec:discussion}
\hspace{4mm}
\textbf{Limitation.}
As the trial to explore conversational human motion generation with visual language models, the proposed MotionChain still has limitations as follows.
MotionChain utilizes indeterministic generative models, similar to other language models, but other traditional or neural motion controllers~\cite{starke2019neural, starke2022deepphase} are mostly deterministic and sensitive to control signals.
Besides, our method can only generate motion on articulated human bodies, excluding many other human parts such as faces~\cite{karras2017audio, cao2018sparse,qiu2022sculptor} and hands~\cite{romero2022embodied, li2022nimble, li2021piano, li2021piano, li2022nimble}.
Although we utilize vision, language, and motion as multimodal conditional inputs akin to human perception, MotionChain is still restricted to the collision signals for human-object and human-scene interactions~\cite{shafir2023priormdm,karunratanakul2023gmd,yuan2023physdiff}.
%

\textbf{Conclusion.}
We summarize the proposed MotionChain as a conversational human motion controller to generate continuous and long-term human motion through multimodal prompts. Compared to these one-turn motion generation methods~\cite{jiang2023motiongpt, zhang2022motiondiffuse, mdm2022human}, our MotionChain produces more contextually rich generation and can achieve the step-by-step process of human task execution for humanoid robotics and game agents.
By leveraging large-scale language, vision-language, and vision-motion data to assist motion-related generation tasks, MotionChain thus comprehends each instruction in multi-turn conversation and generates human motions followed by these prompts.
Extensive experiments validate the efficacy of MotionChain, demonstrating state-of-the-art performance in conversational motion generation, as well as more intuitive manners of controlling and interacting with virtual humans.

\section{Acknowledgment}
This work is supported by National Natural Science Foundation of China (No. 62071127, and 62101137), National Key Research and Development Program of China (No. 2022ZD0160100), Shanghai Natural Science Foundation (No. 23ZR1402900), Shanghai Municipal Science and Technology Major Project (No.2021SHZDZX0103).
The computations in this research were performed using the CFFF platform of Fudan University.

%% file: sections/appendix.tex
\renewcommand\thesection{\Alph{section}}
\renewcommand*{\theHsection}{appedix.\thesection}
\setcounter{section}{0}
\setcounter{figure}{4}
\setcounter{table}{4}
\setcounter{equation}{5}

This appendix provides several additional experiments (\cref{sec:appendix:exps}), more qualitative results (\cref{sec:appendix:qualitative}), 
model implementation details (\cref{sec:appendix:implement}), 
evaluations of inference time (\cref{sec:appendix:inferencetime}),
protocol for the motion conversation evaluation (\cref{sec:appendix:dataset}),
details of motion representations (\cref{sec:appendix:motionRepre}),
metric definitions (\cref{appedix:metrics:details}).



\textbf{Video.} We provide supplemental videos in \href{https://github.com/OpenMotionLab/MotionChain}{Github Page}. In this video, we show 1) examples of motion conversation, 2) comparisons of text-based motion generation, and 3) comparisons of motion reasoning. We suggest the reader watch this video for dynamic motion results.

\textbf{Code} will be available on \href{https://github.com/OpenMotionLab/MotionChain}{GitHub Page}. We provide example code files, which include the process of the training and evaluation of our MotionChain models.

\section{Additional Experiments}
\label{sec:appendix:exps}
We conducted a comprehensive series of experiments to evaluate the efficacy of the proposed MotionChain models further. Specifically, we evaluate each specific comparison on text-to-motion (\cref{appendix:sec:t2m}), motion-to-text (\cref{appendix:sec:m2t}), and motion prediction (\cref{appendix:sec:pred}) on the HumnaML3D~\cite{Guo_2022_CVPR_humanml3d} dataset. Additionally, we present an ablation study focusing on the effectiveness of our motion tokenizer (\cref{sec:vqvae}) and the integration of motion tokens within the language model (\cref{sec:lm}).

\subsection{Comparisons on Text-to-Motion}
\label{appendix:sec:t2m}
The text-to-motion task showcases our MotionGPT model's capability in generating human-like movements based on textual inputs.
Evaluations were performed on MotionChain against current state-of-the-art methods~\cite{chuan2022tm2t,Guo_2022_CVPR_humanml3d,mdm2022human,chen2023mld,zhang2023generating,jiang2023motiongpt}, on the HumanML3D~\cite{Guo_2022_CVPR_humanml3d} dataset according to established metrics~\cite{Guo_2022_CVPR_humanml3d}. The evaluation results, featuring a 95\% confidence interval from 20 runs, largely draw from data reported in the cited works. The comparative outcomes, summarized in \cref{tab:tm:comp:humanml3d}, demonstrating MotionChain's competitive performance across numerous metrics.

\begin{table}[t]
\centering
\vspace{10pt}
\resizebox{\columnwidth}{!}{%
\begin{tabular}{@{}lccccccc@{}}
\toprule
\multirow{2}{*}{Methods}&\multicolumn{3}{c}{RPrecision$\uparrow$}&\multicolumn{1}{c}{\multirow{2}{*}{FID$\downarrow$}}&\multirow{2}{*}{MMDist$\downarrow$}&\multirow{2}{*}{Diversity$\rightarrow$}&\multirow{2}{*}{MModality$\uparrow$}\\\cmidrule(lr){2-4}
&\multicolumn{1}{c}{Top1}&\multicolumn{1}{c}{Top2}&\multicolumn{1}{c}{Top3}&\multicolumn{1}{c}{}&&&\\\midrule
Real &
  $0.511^{\pm.003}$ &
  $0.703^{\pm.003}$ &
  $0.797^{\pm.002}$ &
  $0.002^{\pm.000}$ &
  $2.974^{\pm.008}$ &
  $9.503^{\pm.065}$ &
  \multicolumn{1}{c}{-}
  \\ \midrule
TM2T \cite{chuan2022tm2t} &
  $0.424^{\pm.003}$ &
  $0.618^{\pm.003}$ &
  $0.729^{\pm.002}$ &
  $1.501^{\pm.017}$ &
  $3.467^{\pm.011}$ &
  $8.589^{\pm.076}$ &
  $\underline{2.424}^{\pm.093}$ 
  \\
T2M \cite{Guo_2022_CVPR_humanml3d}&
  $0.457^{\pm.002}$ &
  $0.639^{\pm.003}$ &
  $0.740^{\pm.003}$ &
  $1.067^{\pm.002}$ &
  $3.340^{\pm.008}$ &
  $9.188^{\pm.002}$ &
  $2.090^{\pm.083}$ \\
MotionDiffuse \cite{zhang2022motiondiffuse} &
$\underline{0.491}^{\pm.001}$ &
$\underline{0.681}^{\pm.001}$ &
$\underline{0.782}^{\pm.001}$ &
$0.630^{\pm.001}$ &
${3.113}^{\pm.001}$ &
${9.410}^{\pm.049}$ &
$1.553^{\pm.042}$ \\
MDM \cite{mdm2022human}&
  $0.320^{\pm.005}$ &
  $0.498^{\pm.004}$ &
  $0.611^{\pm.007}$ &
  ${0.544}^{\pm.044}$ &
  $5.566^{\pm.027}$ &
  ${9.559}^{\pm.086}$ &
  $\underline{2.799}^{\pm.072}$ \\
MLD \cite{chen2023mld}&
  ${0.481}^{\pm.003}$ &
  ${0.673}^{\pm.003}$ &
  ${0.772}^{\pm.002}$ &
  ${0.473}^{\pm.013}$ &
  ${3.196}^{\pm.010}$ &
  $9.724^{\pm.082}$ &
  ${2.413}^{\pm.079}$
  \\
T2M-GPT \cite{zhang2023generating}  & $\underline{0.491}^{\pm.003}$ & ${0.680}^{\pm.003}$ & ${0.775}^{\pm.002}$ & $\boldsymbol{0.116}^{\pm.004}$ & ${3.118}^{\pm.011}$ & $9.761^{\pm.081}$ & $1.856^{\pm.011}$ \\
MotionGPT \cite{jiang2023motiongpt} &
${0.492}^{\pm.003}$ & $\underline{0.681}^{\pm.003}$ & ${0.778}^{\pm.002}$ & $\underline{0.232}^{\pm.008}$ & ${3.096}^{\pm.008}$ & $\boldsymbol{9.528}^{\pm.071}$ & 
${2.008}^{\pm.084}$
\\
\midrule
MotionChain &
$\boldsymbol{0.504}^{\pm.003}$ & $\boldsymbol{0.695}^{\pm.003}$ &	$\boldsymbol{0.790}^{\pm.003}$ &${0.248}^{\pm.009}$ &	$\boldsymbol{3.033}^{\pm.010}$ &$\underline{9.470}^{\pm.075}$ &	${1.715}^{\pm.066}$
\\ \bottomrule
\end{tabular}%
}
\vspace{5pt}
\caption{Comparison of text-to-motion on HumanML3D~\cite{Guo_2022_CVPR_humanml3d}. The empty MModality indicates \textit{Real} motion is deterministic. \textit{Pre-trained} and \textit{Fine-tuned} indicate uniform motion-language pre-training and specific fine-tuning on this task. The arrows ($\rightarrow$) indicate that closer to \textit{Real} is desirable. \textbf{Bold} and \underline{underline} indicate the best and the second best result on text-to-motion task.}
\vspace{-5pt}
\label{tab:tm:comp:humanml3d}
\end{table}

\subsection{Comparisons on Motion-to-Text}
\label{appendix:sec:m2t}
In the motion-to-text task, the goal is to generate descriptive text based on sequences of human motion. We evaluate the proposed MotionChain, contrasting it with TM2T~\cite{chuan2022tm2t} and MotionGPT~\cite{jiang2023motiongpt} on the HumanML3D dataset and adhering to the evaluation metrics used in ~\cite{chuan2022tm2t,jiang2023motiongpt}. Following ~\cite{jiang2023motiongpt}, we leverages the original ground truth texts for evaluation, ensuring a more comprehensive assessment . Assessments in \cref{tab:tm:comp:m2t} demonstrate that MotionChain outperforms the recent methods in generating text descriptions of human motions on most benchmarks.

\begin{table}[t]
\resizebox{\columnwidth}{!}{%
\begin{tabular}{@{}lcccccc@{}}
\toprule
\multirow{2}{*}{Methods}
&\multirow{2}{*}{$\text{Length}_{\text{avg}}$$\uparrow$} &\multirow{2}{*}
{Bleu@1$\uparrow$}&
\multirow{2}{*}{Bleu@4$\uparrow$}&\multirow{2}{*}{Rouge$\uparrow$}&\multirow{2}{*}{Cider$\uparrow$}&\multirow{2}{*}{BertScore$\uparrow$}
\\
\\\midrule

Real 
&  $12.75$ &\multicolumn{1}{c}{-}& \multicolumn{1}{c}{-}& \multicolumn{1}{c}{-}& \multicolumn{1}{c}{-}& \multicolumn{1}{c}{-}
\\ \midrule

TM2T\cite{chuan2022tm2t} 
& $10.67$& $\boldsymbol{48.9}$ & $7.00$ & ${38.1}$ & $16.8$ & ${32.2}$ \\
MotionGPT~\cite{jiang2023motiongpt}
& $\boldsymbol{13.04}$ 
& $48.2$ & ${12.47}$ & $37.4$ & ${29.2}$ & ${32.4}$
\\ \midrule
MotionChain
& ${12.37}$ 
& $48.1$ & $\boldsymbol{12.56}$ & $\boldsymbol{39.9}$ & $\boldsymbol{33.7}$ & $\boldsymbol{36.9}$

\\
   \bottomrule
\end{tabular}%
}
\vspace{5pt}
\caption{Comparison of motion captioning on HumanML3D~\cite{Guo_2022_CVPR_humanml3d}. The evaluation metrics follow \cite{chuan2022tm2t}, while we use the ground truth texts without pre-processing for linguistic metrics calculation. \textbf{Bold} indicate the best.}
\vspace{-15pt}
\label{tab:tm:comp:m2t}
\end{table}

\subsection{Comparisons on Motion Completion.}
\label{appendix:sec:pred}
In accordance with MotionGPT~\cite{jiang2023motiongpt}, we consider motion prediction as a collective task referred to as general motion completion. To assess the motion completion capability of MotionChain, we utilize a subset of the AMASS dataset~\cite{AMASS_ICCV2019}, which consists solely of motion data. For the motion prediction task, we use only the initial 20\% of the motion sequence as conditions. We evaluate MotionChain using the identical settings as outlined in ~\cite{jiang2023motiongpt}. The motion completion results of MotionChain, presented in Table \ref{tab:comp:motion}, indicate that MotionChain achieves lower values in terms of ADE and FDE metrics. This implies that the mean and last-frame L2 distance between the ground truth and predicted motion are closer.

\begin{table}[h]
\centering
\resizebox{0.7\columnwidth}{!}{%
\begin{tabular}{@{}lccccccccc@{}}
\toprule
\multirow{2}{*}{Methods} & 
\multicolumn{4}{c}{Motion Prediction }
\\
\cmidrule(lr){2-5} 
& $\text{FID}\downarrow $  
& Diversity$\uparrow$            
& ADE$\downarrow$
& FDE$\downarrow$
\\ 
\toprule
Real & $0.002$ & $9.503$ & - & - 
\\\midrule
MDM\cite{mdm2022human}& $6.031$ & $7.813$& $5.446$ &$8.561$
\\
T2M-GPT\cite{zhang2023generating}& $2.056$ & $8.635$& $6.161$& $8.302$ 
\\
MotionGPT~\cite{jiang2023motiongpt}
& $\boldsymbol{0.905}$ &$\boldsymbol{8.972}$ & ${4.745}$& ${6.040}$ 
\\ \midrule
MotionChain
& ${1.053}$ &${8.802}$ & $\boldsymbol{4.388}$& $\boldsymbol{5.401}$ 
\\\bottomrule
\end{tabular}%
}
\vspace{5pt}
\caption{Comparison of motion prediction and motion in-between on part of AMASSS~\cite{AMASS_ICCV2019} dataset using motion data only. 
FID indicates motion quality and Diversity (DIV) for motion diversity within each condition. ADE and FDE are joints distance between generation and ground truth.}
\vspace{-15pt}
\label{tab:comp:motion}
\end{table}

\subsection{Ablation on Motion Tokenizer.} 
\label{sec:vqvae}
we conducted an ablation study on the motion tokenizer $\mathcal{V}$ of the MotionChain model, focusing specifically on the impact of varying the size $K$ and dimension $d$ of motion codebooks, and residual quantizer layers $Q$. Additionally, we benchmarked our VQ-VAE implementation against previous work~\cite{vposer_SMPL-X:2019, petrovich21actor, chen2023mld}, as shown in \cref{tab:mr:ablation}. This comparative analysis underscored the better performance of our VQ-VAE approach in terms of motion reconstruction accuracy. Through this comprehensive ablation study, in addition to the length limit of T5 series models, we thus identified parameters for the majority of our experiments as $Q=4, K=512, d=1024$.

\begin{table}[t]
\centering
\vspace{10pt}
\resizebox{\columnwidth}{!}{%
\begin{tabular}{@{}lcccccccc@{}}
\toprule
\multirow{2}{*}{Methods}&\multirow{2}{*}{Motion Token Numbers}&\multicolumn{3}{c}{RPrecision$\uparrow$}&\multicolumn{1}{c}{\multirow{2}{*}{FID$\downarrow$}}&\multirow{2}{*}{MMDist$\downarrow$}&\multirow{2}{*}{Diversity$\rightarrow$}&\multirow{2}{*}{MModality$\uparrow$}\\\cmidrule(lr){3-5}
&&\multicolumn{1}{c}{Top1}&\multicolumn{1}{c}{Top2}&\multicolumn{1}{c}{Top3}&\multicolumn{1}{c}{}&&&\\\midrule
Real & - &
  $0.511^{\pm.003}$ &
  $0.703^{\pm.003}$ &
  $0.797^{\pm.002}$ &
  $0.002^{\pm.000}$ &
  $2.974^{\pm.008}$ &
  $9.503^{\pm.065}$ &
  \multicolumn{1}{c}{-}
  \\
\midrule
Shared & $V_m$ &
$0.496^{\pm.003}$ & $0.686^{\pm.003}$ & $0.784^{\pm.002}$ & $0.291^{\pm.012}$ & $3.067^{\pm.011}$ & $9.394^{\pm.075}$ & $\boldsymbol{2.072}^{\pm.080}$
\\
Independent &$V_m\times Q$ &
$\boldsymbol{0.504}^{\pm.003}$ & $\boldsymbol{0.695}^{\pm.003}$ &	$\boldsymbol{0.790}^{\pm.003}$ &$\boldsymbol{0.248}^{\pm.009}$ &	$\boldsymbol{3.033}^{\pm.010}$ &$\boldsymbol{9.470}^{\pm.075}$ &	${1.715}^{\pm.066}$
\\ \bottomrule
\end{tabular}%
}
\vspace{5pt}
\caption{Comparison of text-to-motion on HumanML3D~\cite{Guo_2022_CVPR_humanml3d}. The empty MModality indicates \textit{Real} motion is deterministic. \textit{Pre-trained} and \textit{Fine-tuned} indicate uniform motion-language pre-training and specific fine-tuning on this task. The arrows ($\rightarrow$) indicate that closer to \textit{Real} is desirable. \textbf{Bold} and \underline{underline} indicate the best and the second best result on text-to-motion task.}
\vspace{-5pt}
\label{tab:abl:share}
\end{table}

\begin{table}[t]
\centering
\resizebox{0.8\columnwidth}{!}{
\begin{tabular}{@{}lcccc@{}}
\toprule
\multirow{2}{*}{Method}&\multicolumn{4}{c}{Reconstruction}
\\\cmidrule(lr){2-5}
&MPJPE$\downarrow$&PAMPJPE$\downarrow$

&FID$\downarrow$&DIV$\rightarrow$
\\ \midrule
Real  &  \multicolumn{1}{c}{-} & \multicolumn{1}{c}{-}

&  $0.002$&	$9.503$     \\ \midrule
VPoser-t~\cite{vposer_SMPL-X:2019}  & $75.6$ & $48.6$

&     $1.430$ & $8.336$   \\
ACTOR~\cite{petrovich21actor}     & $65.3$ & $41.0$

&    $0.341$ & $\boldsymbol{9.569}$ \\
MLD-1~\cite{chen2023mld}  & $\boldsymbol{54.4}$ & $41.6$ 

& $0.247$ & $9.630$ \\
MotionGPT~\cite{jiang2023motiongpt} & 
$55.8$ &$\boldsymbol{40.1}$  

&${0.067}$&$9.675$	\\
 \midrule
MotionChain & ${63.1}$ &${43.4}$  

&$\boldsymbol{0.014}$&${9.157}$
\\ \bottomrule
\toprule
$Q=4, K=128, d=512$ & $71.8$ & $ 51.2$ & $0.037$ & $9.098$\\
$Q=4, K=256, d=512$& $70.4$ & $ 48.5$ & $0.051$ & $9.004$\\
$Q=4, K=512, d=512$ & $69.5$ & $ 46.5$ & $\boldsymbol{0.025}$ & $9.015$
\\
$Q=4, K=1024, d=512$ & $\boldsymbol{65.9}$ & $ \boldsymbol{43.9}$ & $0.041$ & $\boldsymbol{9.310}$
\\\midrule
$Q=2, K=512, d=512$ & $79.7$ & $ 56.9$& $0.081$ & $9.162$\\
$Q=4, K=512, d=512$ & $69.5$ & $ 46.5$ & $0.025$ & $9.015$	\\
$Q=8, K=512, d=512$ &  $49.7$ & $ 38.6$ & $\boldsymbol{0.025}$ & $\boldsymbol{9.213}$	\\
$Q=16, K=512, d=512$ & $\boldsymbol{48.4}$ &$\boldsymbol{38.4}$ 	&${0.026}$&${9.075}$
\\\midrule
$Q=4, K=512, d=128$ & ${114.5}$ &${79.7}$ 	&${1.698}$&$8.344$ 
\\
$Q=4, K=512, d=256$ & $83.9$ & $ 59.7$& $0.560$ & $8.782$
\\
$Q=4, K=512, d=512$ &  $69.5$ & $ 46.5$& $0.052$ & $9.015$
\\
$Q=4, K=512, d=1024$ & $\boldsymbol{63.1}$ &$\boldsymbol{43.4}$ 	&$\boldsymbol{0.014}$&$\boldsymbol{9.157}$
\\
\bottomrule
\end{tabular}
}
\vspace{10pt}
\caption{Evaluation of our motion tokenizer on the motion part of HumanML3D~\cite{Guo_2022_CVPR_humanml3d} dataset. We follow MLD~\cite{chen2023mld} to evaluate our VQ-VAE model $\mathcal{V}$: MPJPE and PAMPJPE are measured in millimeter. ACCL indicates acceleration error. We evaluate FID and Diversity the same as Tab. 3. The baselines of VPoser-t~\cite{vposer_SMPL-X:2019} and ACTOR~\cite{petrovich21actor} are borrowed from MLD. $K$ indicates the codebook size, $d$ indicates the codebook dimension , $Q$ indicates the Residual-VQ layers.}
\label{tab:mr:ablation}
\end{table}

\subsection{Ablation on Motion Tokens.} 
\label{sec:lm}
Subsequent to our analysis of motion codebooks, we shift focus to the strategy of sharing motion vocabularies $V_m$ within the language model backbone. Specifically, we examine the efficacy of sharing motion codes across various residual quantization layers versus a more isolated approach. This means a comparison between integrating 
$V\times Q$ newly added tokens and $V$ tokens alone within the language models. Our experiment shown in \cref{tab:abl:share}, grounded in the text-to-motion experiments conducted on the HumanML3D~\cite{Guo_2022_CVPR_humanml3d} dataset, reveals that the best performance is achieved when motion codes are not shared across the language model.

\newpage

\section{Qualitative Results}
\label{sec:appendix:qualitative}
\begin{figure}[H]
    \includegraphics[width=\linewidth]{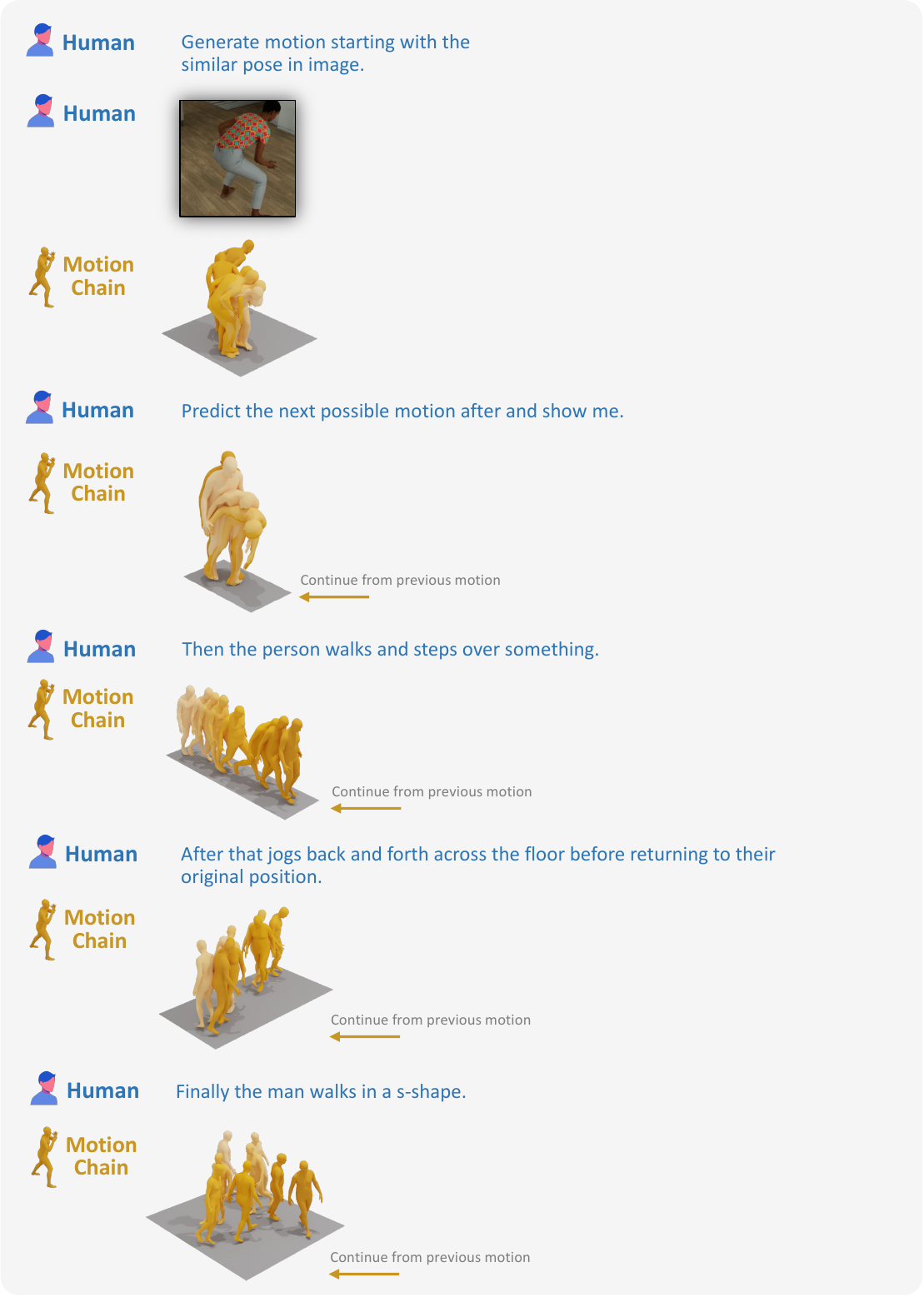}
    \caption{
    The gallery showcases the results of our MotionChain model. The supervision of MotionChain is based on our conversational motion-language dataset (see Appendix~\ref{sec:appendix:dataset}), which builds upon previous motion datasets~\cite{Guo_2022_CVPR_humanml3d,BABEL:CVPR:2021}. For a more dynamic visualization, we recommend referring to our supplemental video.}
    \label{fig:appendix:gallery}
\end{figure}

\begin{figure}[H]
    \includegraphics[width=\linewidth]{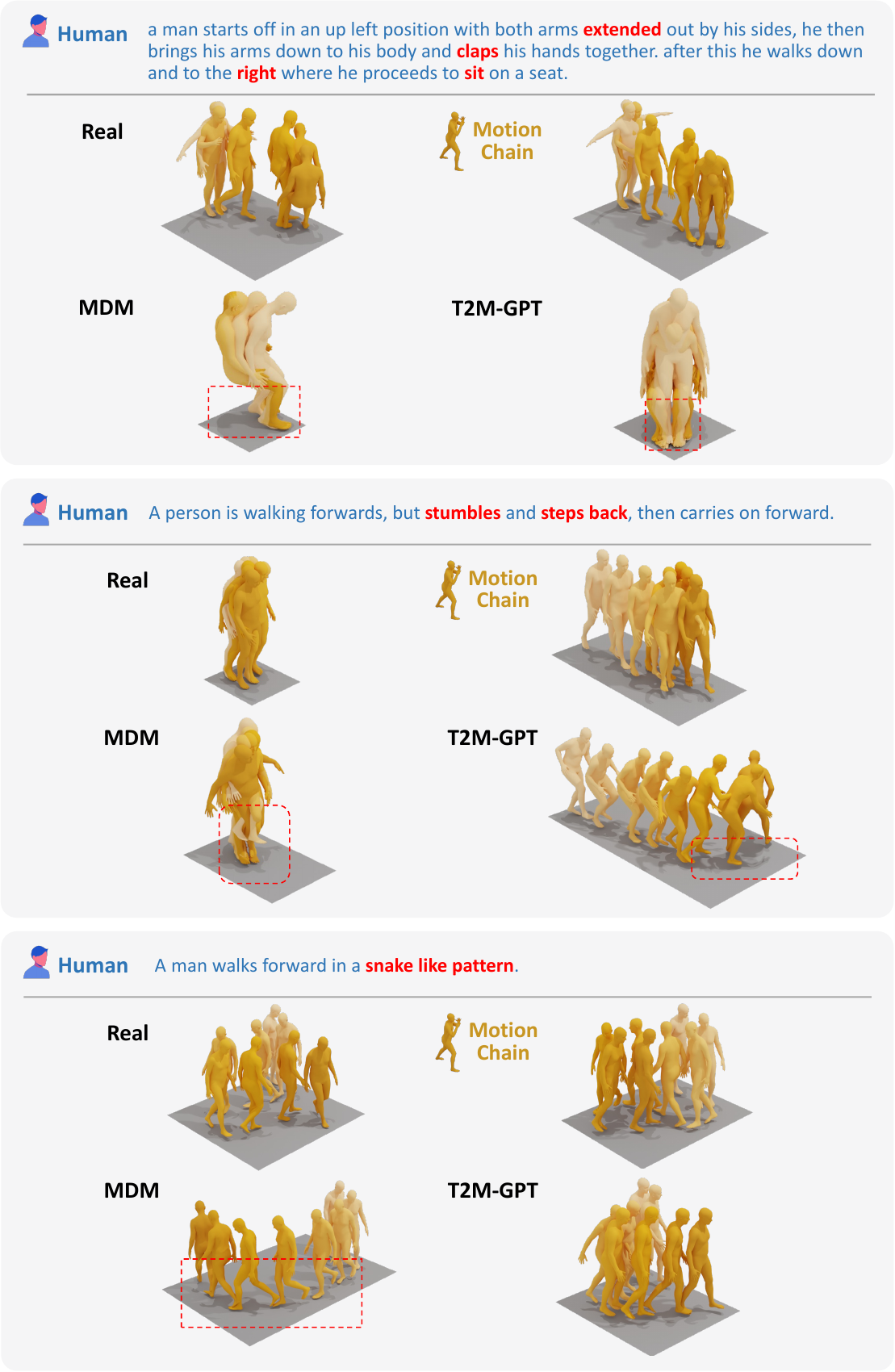}
    \caption{Comparison of text-driven motion generation methods on the HumanML3D dataset~\cite{Guo_2022_CVPR_humanml3d}. In the visualizations, misaligned motions are highlighted with red words and boxes, while the characters are color-coded from light to dark to indicate the progression of time.}
    \label{fig:appendix:compt2m}
\end{figure}

\begin{figure}[H]
    \includegraphics[width=\linewidth]{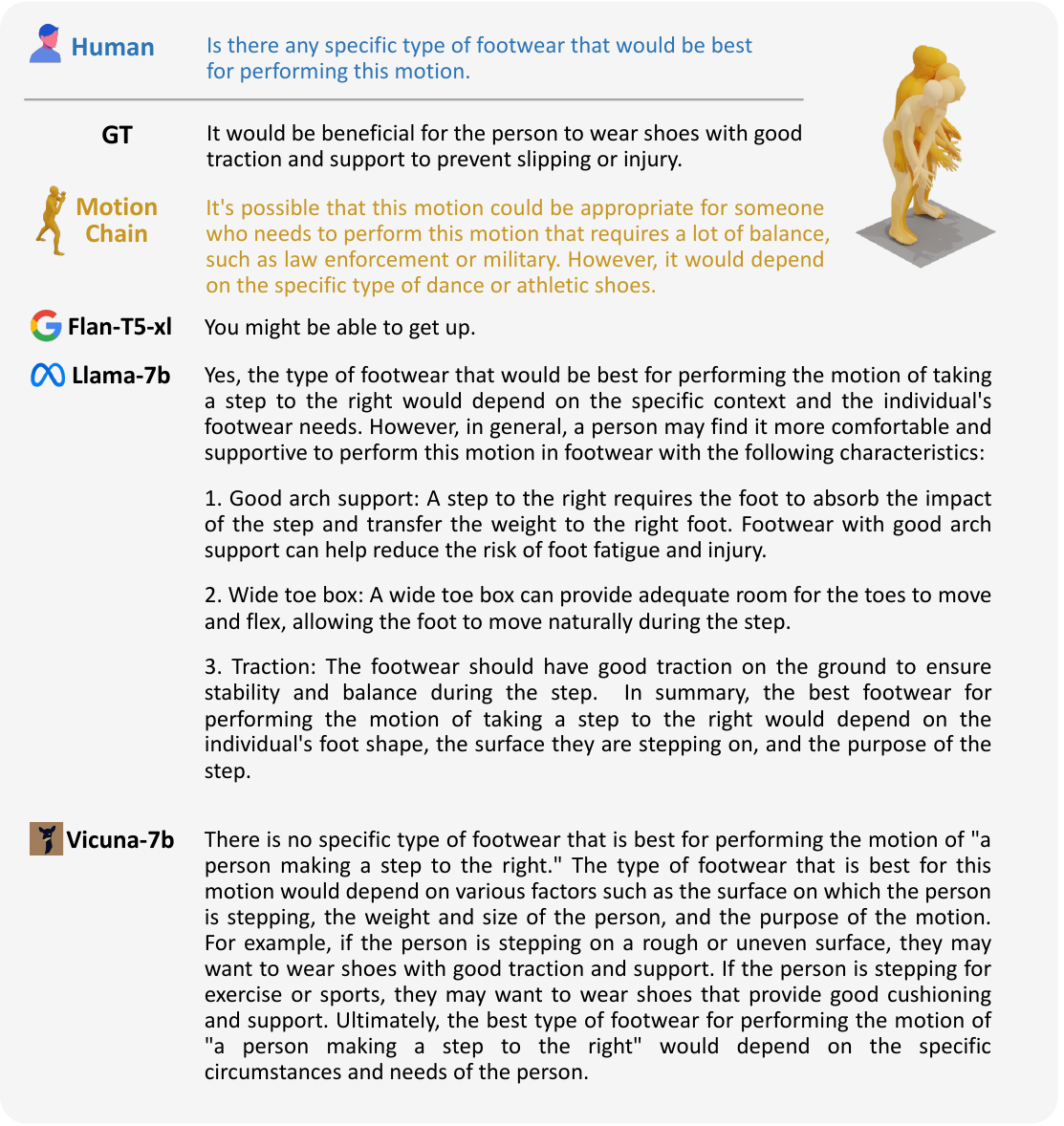}
    
    \caption{Comparison on motion reasoning question-answer. The MotionChain is trained on our conversation dataset based on HumanML3D~\cite{Guo_2022_CVPR_humanml3d}. The results demonstrate that our MotionChain shows promising text and motion understanding.}
    \label{fig:appendix:conv}
\end{figure}

We visualize our result gallery on motion conversations ($cf.$ \cref{fig:appendix:gallery}) and some qualitative results on the comparison of text-to-motion ($cf.$ \cref{fig:appendix:compt2m}) and motion reasoning ($cf.$ \cref{fig:appendix:conv}).
\newpage

\begin{figure}[H]
    \includegraphics[width=\linewidth]{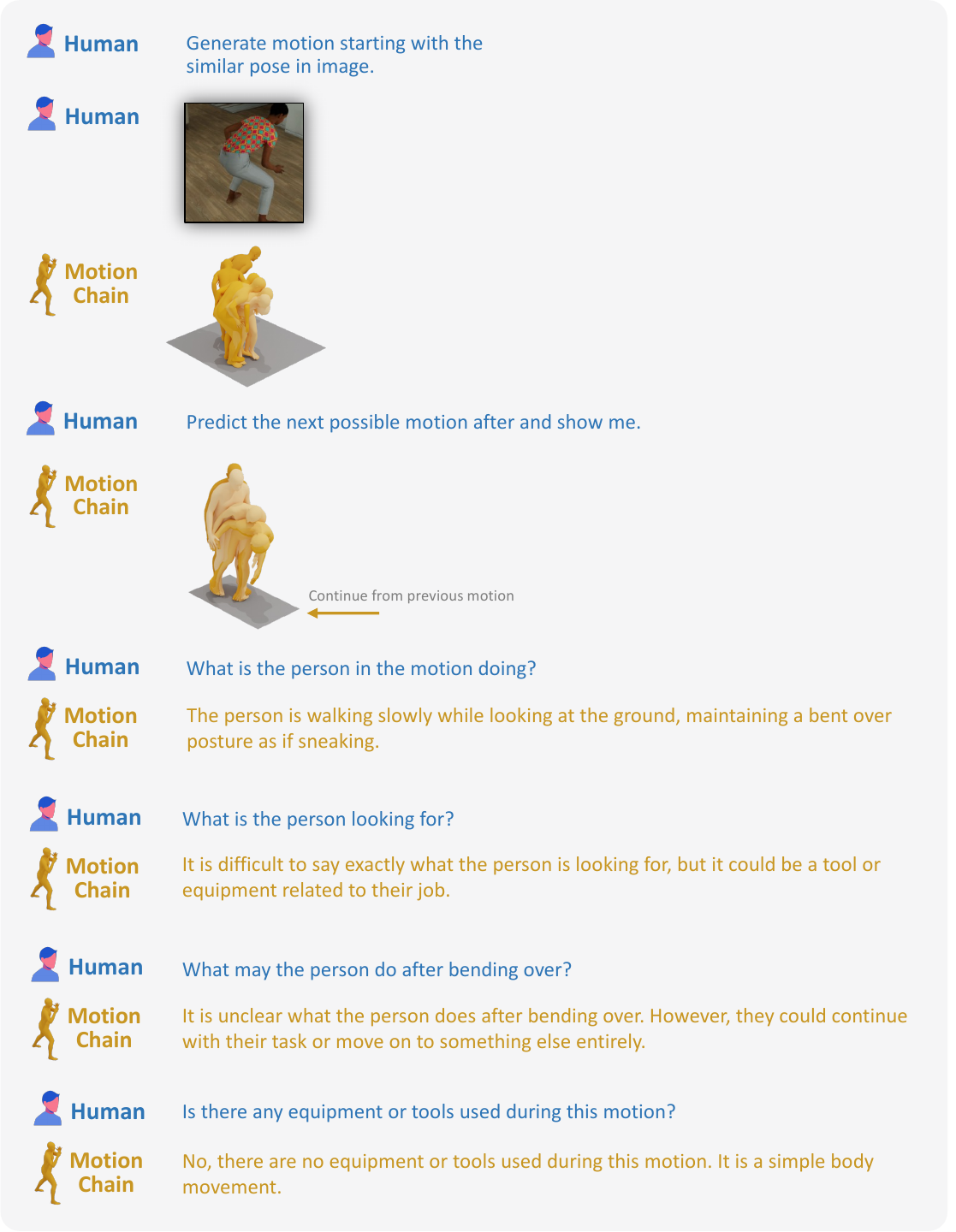}
    \centering
\end{figure}

\begin{figure}[H]
    \includegraphics[width=\linewidth]{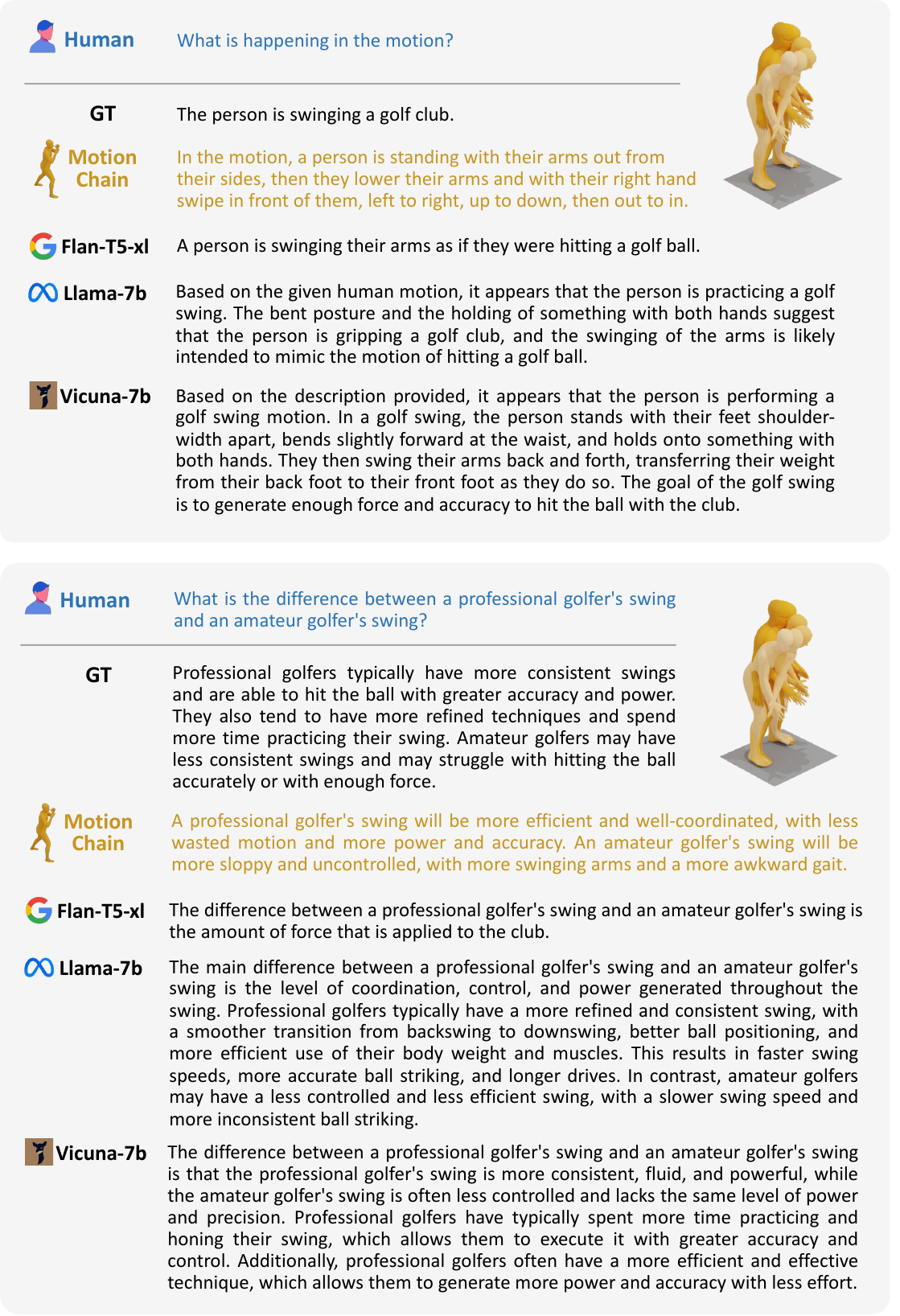}
    \centering
\end{figure}

\begin{figure}[H]
    \includegraphics[width=\linewidth]{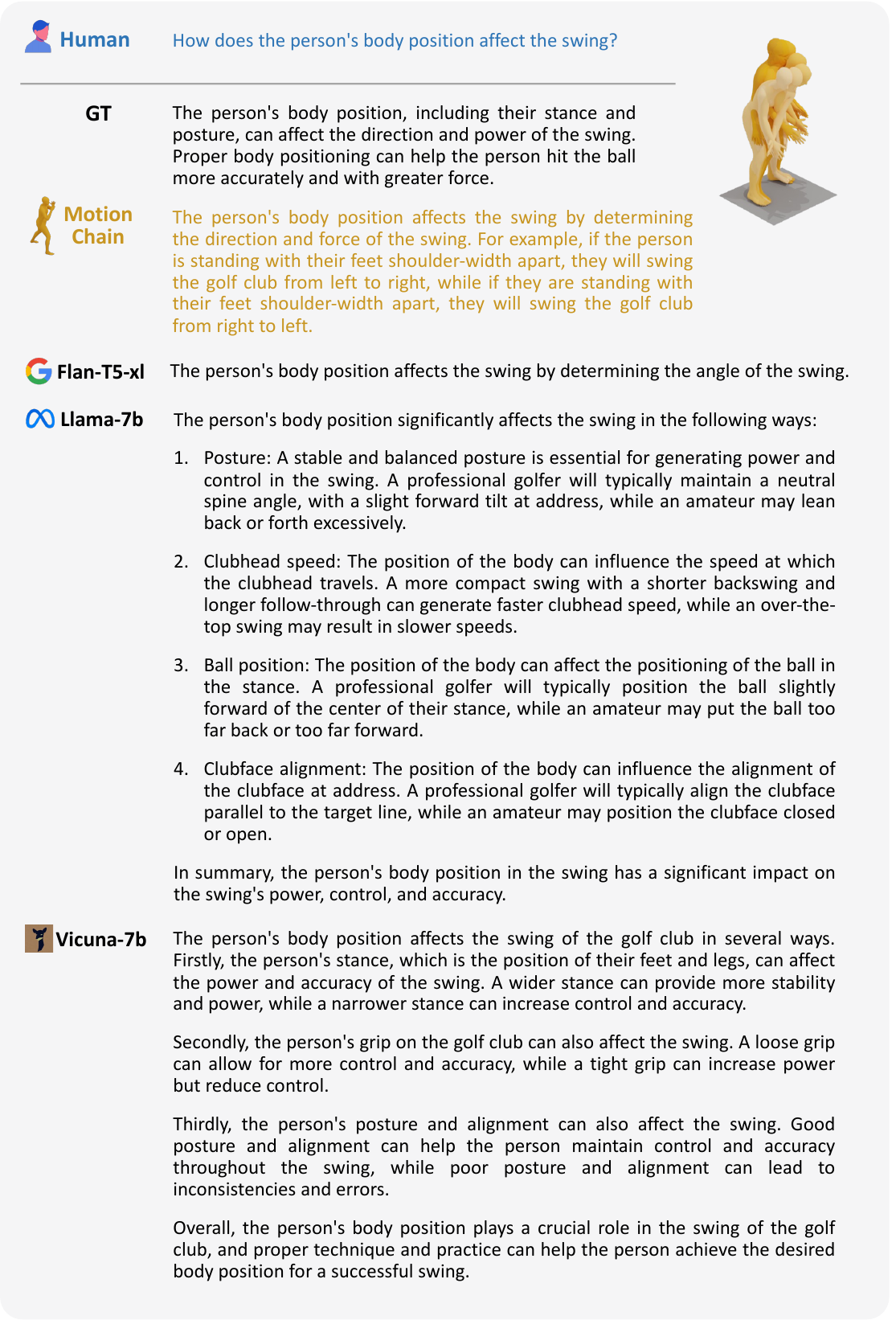}
    \centering
\end{figure}

\begin{figure}[H]
    \includegraphics[width=\linewidth]{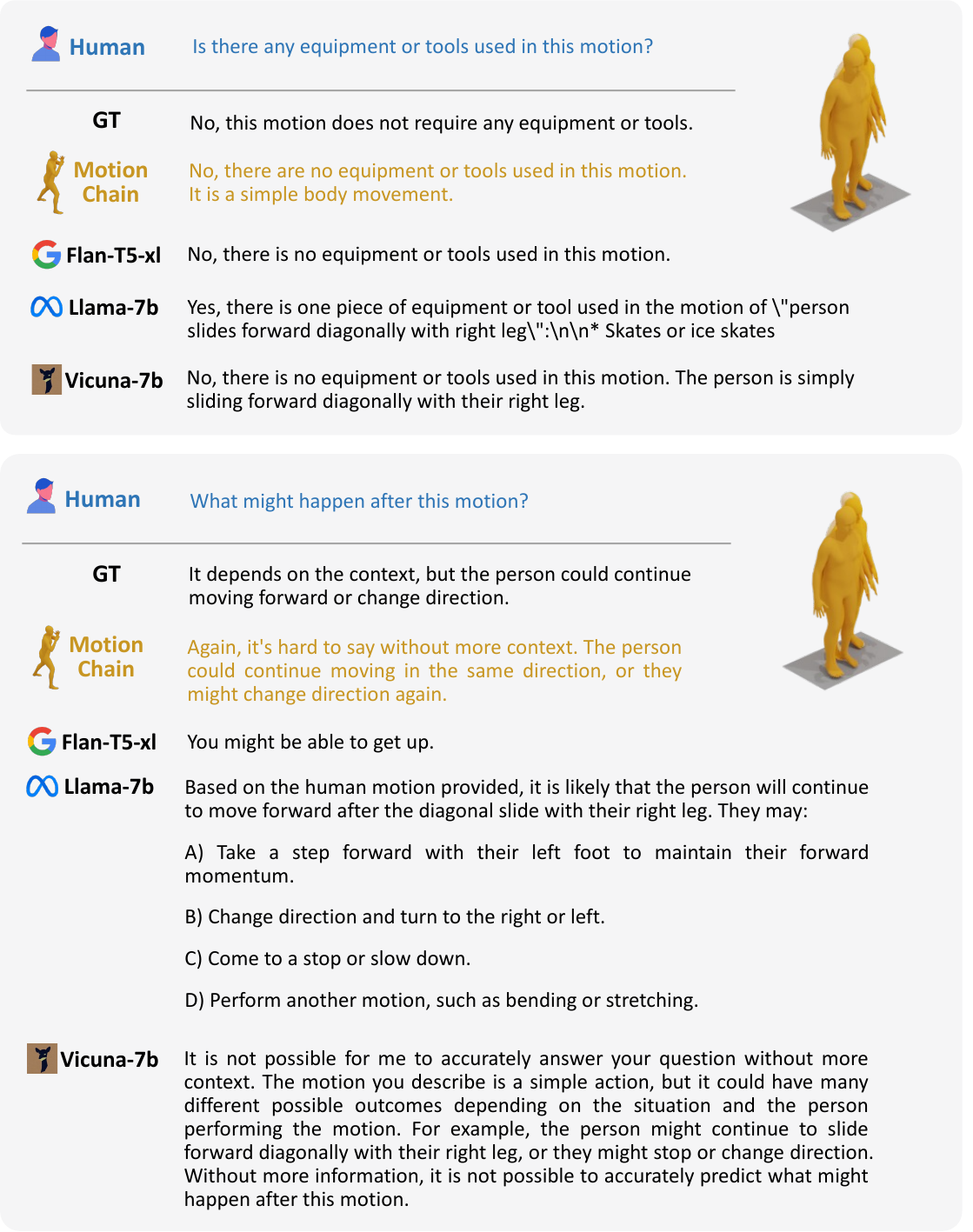}
    \centering
\end{figure}

\section{Implementation Details}
\label{sec:appendix:implement}

We provide detailed explanations regarding the implementation details of motion composition ( \cref{appendix:detail:composition}), and the image tokenizer ( \cref{sec:appendix:detail:image}).

\subsection{Details of Temporal Motion Compoistion}
\label{appendix:detail:composition}
To investigate the temporal motion composition abilities of the MotionChain model, we conduct a pair actions composition experiment on the BABEL dataset~\cite{BABEL:CVPR:2021}, following the methodology of TEACH~\cite{TEACH:3DV:2022}. For simplicity, we consider pairs of actions, but it is important to note that MotionChain can handle sequences of actions/motions of any length. During training, in cases where there is segment overlap, we evenly distribute the overlapping frames between the two segments that form the pair. It is worth mentioning that the majority of the pair data (approximately 70 \% ) is generated through overlapping segments rather than transitions. In the event of a transition, we concatenate the transition with the second segment.
Instead of training a MotionChain model from scratch on the BABEL dataset~\cite{BABEL:CVPR:2021}, we utilize a pre-trained MotionChain model obtained from HumanML3D~\cite{Guo_2022_CVPR_humanml3d}. Subsequently, we convert the motion data in the BABEL dataset~\cite{BABEL:CVPR:2021} into the format used in HumanML3D~\cite{Guo_2022_CVPR_humanml3d}, and then fine-tune the MotionChain model on the BABEL dataset~\cite{BABEL:CVPR:2021} using prompts that incorporate memory, as demonstrated below:
\begin{quote}

\small$X_{\text{system-message}}$ 

USER: Please assume the role of an Human Motion Language translator. I will use English, you should translate it, and respond in Human Motion Language. My first request is "\textcolor{Blue}{$<$label1$>$}" 

ASSISTANT: \textcolor{Orange}{$<$motion1$>$}

USER: Please assume the role of a Human Motion Language translator. I will use English, you should translate it, and respond in Human Motion Language. In the last round I asked you to translate "\textcolor{Blue}{$<$label1$>$}", and your answer is \textcolor{Orange}{$<$motion1$>$}. Now my second request is "\textcolor{Blue}{$<$label2$>$}" 

ASSISTANT: \textcolor{Orange}{$<$motion2$>$}

\end{quote}
For comparison with TEACH~\cite{TEACH:3DV:2022}, we employed the TEACH model that was pre-trained on the BABEL dataset~\cite{BABEL:CVPR:2021} to generate motion samples 20 times on the validation set. Subsequently, we converted the generated motion, originally in SMPL-H format~\cite{SMPL2015}, into the HumanML3D format.

We also examine the influence of various motion composition mechanisms on the generated complete motion sequences, as presented in Table 3. The "Independent" mechanism refers to the direct concatenation of independently generated motion sequences without any additional processing. On the other hand, the "Tokens-joint" mechanism involves concatenating motion tokens and decoding them using the VQ decoder, which results in a more coherent and natural sequence of movements.

\subsection{Details of Image Tokenzier}
\label{sec:appendix:detail:image}
We explore three different architectural designs for image tokenizers:

(a) \textit{MLP}: In this design, we connect the frozen vision encoder CLIP ViT-L/14 \cite{radford2021clip} to the language model using a linear layer. The output of the vision encoder is projected to the same dimension as the word embeddings of the language model and is inserted before the text or motion token embeddings.

(b) \textit{Perceiver}: This design incorporates a perceiver module with a similar architecture to Flamingo~\cite{alayrac2022flamingo}. The perceiver module includes a transformer that receives a predefined number of latent input queries. These queries are then projected to the same dimension as the word embeddings of the language model and are inserted before the text or motion token embeddings. Details of architecture is presented in \cref{tab:appendix:perceiver}.

(c) \textit{Q-former}: In this design, we directly utilize the pre-trained Q-former from BLIP-2~\cite{li2023blip2} to align visual inputs with the language model. The Q-former is frozen throughout the entire training process.

\begin{table}[]
\begin{tabular}{l}
\toprule
(0): PerceiverResampler( \\
\quad(layers): ModuleList( \\
\quad\quad(0-5): 6 x ModuleList( \\
\quad\quad\quad(0): PerceiverAttention( \\
\quad\quad\quad\quad(norm\_media): LayerNorm((1024,), eps=1e-05, elementwise\_affine=True) \\
\quad\quad\quad\quad(norm\_latents): LayerNorm((1024,), eps=1e-05, elementwise\_affine=True) \\
\quad\quad\quad\quad (to\_q): Linear(in\_features=1024, out\_features=512, bias=False) \\
\quad\quad\quad\quad (to\_kv): Linear(in\_features=1024, out\_features=1024, bias=False) \\
\quad\quad\quad\quad (to\_out): Linear(in\_features=512, 
 out\_features=1024, bias=False) ) \\
\quad\quad\quad (1): Sequential( \\
\quad\quad\quad\quad (0): LayerNorm((1024,), eps=1e-05, elementwise\_affine=True) \\
\quad\quad\quad\quad (1): Linear(in\_features=1024, out\_features=4096, bias=False) \\
\quad\quad\quad\quad (2): GELU(approximate='none') \\
\quad\quad\quad\quad (3): Linear(in\_features=4096, out\_features=1024, bias=False) ) ) ) \\
\quad(norm): LayerNorm((1024,), eps=1e-05, elementwise\_affine=True) ) \\
(1): Linear(in\_features=1024, out\_features=768, bias=True)       
\\ \bottomrule
\end{tabular}
\vspace{5pt}
\caption{Architecture of our vision perceiver}
\label{tab:appendix:perceiver}
\vspace{-10pt}
\end{table}

\section{Inference time}
\label{sec:appendix:inferencetime}
We conducted a study to evaluate the inference time of our MotionChain model, which utilizes an auto-regressive approach for motion generation. To assess the time costs, we measured the Frames Per Second (FPS) on a single Tesla V100 GPU with a batch size of one. It is important to note that the frame generation rate of MotionChain, even without specific engineering optimizations, surpasses the ground-truth frame rate in text-motion pair datasets~\cite{Guo_2022_CVPR_humanml3d,BABEL:CVPR:2021,lin2023motionx}, highlighting its capability to support real-time motion animation applications.

\begin{table}[h]
\centering
\resizebox{0.5\columnwidth}{!}
{%
\begin{tabular}{@{}llcc@{}}
\toprule
Models & Backbone &Parameters
& FPS
\\ \midrule
MotionChain-small & Flan-T5-small & 110 M & 136.7
\\
MotionChain-base & Flan-T5-Base & 280 M & 74.99
\\
MotionChain-large & Flan-T5-Large & 810 M & 39.18
\\
\bottomrule
\end{tabular}%
}
\vspace{5pt}
\caption{The inference time costs of text-driven motion generation by evaluating the Frames Per Second (FPS), which is obtained by averaging the number of frames generated per second. We present the time costs for various model sizes and observe that, under the same 1 Tesla V100, smaller model sizes achieve faster FPS.}
\label{tab:tm:abl:scheduler}
\vspace{-10pt}
\end{table}

\section{Evaluation Protocols on the Motion Conversation.}
\label{sec:appendix:dataset}
\begin{figure}[h]
    \centering
    \includegraphics[width=\linewidth]{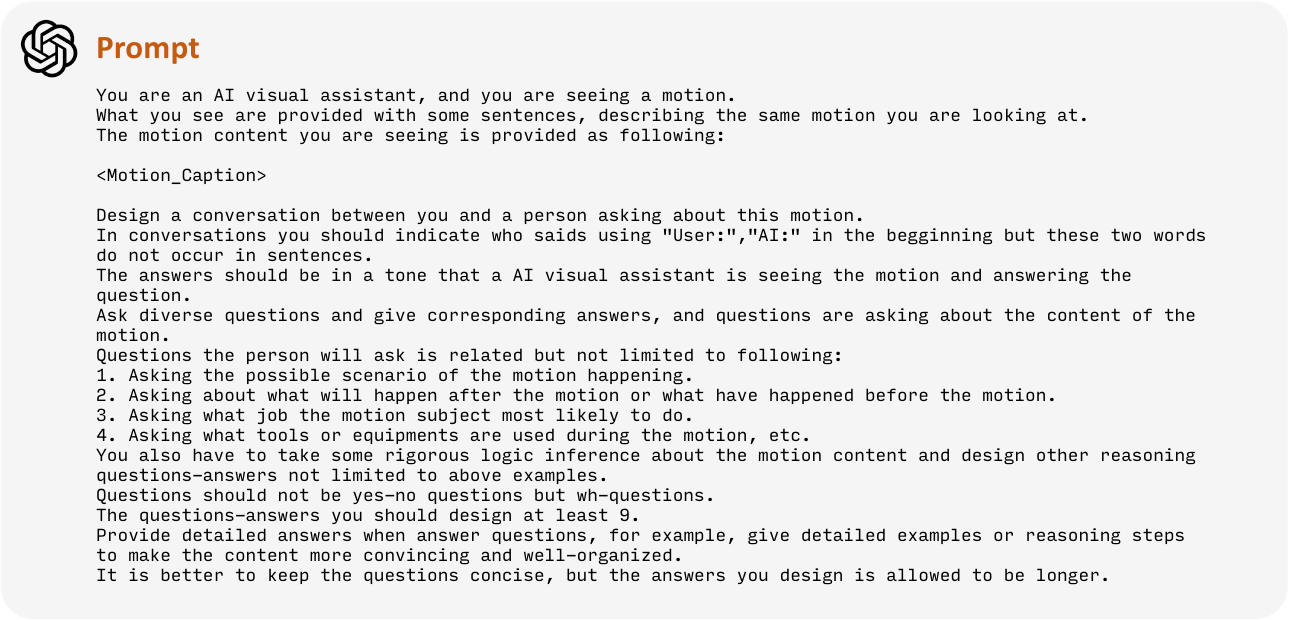}
    \caption{
    The dedicated ChatGPT prompt for facilitating the collection of motion question-answer pairs. Our primary goal was to encompass a wide range of topics, including motion physics and motion analysis. By utilizing this prompt, our aim was to enable ChatGPT to generate high-quality questions, thereby making a valuable contribution to the development of a comprehensive motion question-answer dataset.
    }
    \label{fig:appendix:prompt}
\end{figure}

We propose a protocol to evaluate our Multi-turn Multi-modal model, MotionChain, on various motion-language generation tasks. While MotionGPT~\cite{jiang2023motiongpt} utilized previous text-motion pair datasets~\cite{Guo_2022_CVPR_humanml3d,Plappert2016kit,AMASS_ICCV2019} to create an instruction motion-language dataset comprising 14 core tasks with numerous instruction templates, these tasks lack analysis of human motion and are limited to single-turn generation without contextual memory. To overcome this limitation, we introduce motion reasoning and motion editing tasks that leverage contextual information. Initially, we manually provide ChatGPT~\cite{ouyang2022instructgpt,openai2023gpt4} with a few examples along with corresponding textual descriptions of the motions in the datasets, and then we let it generate the motion analysis (refer to ~\cref{fig:appendix:prompt}). Additionally, using a pre-trained text-motion retrieval model, TMR~\cite{petrovich2023tmr}, we retrieve motions from the dataset with high and middle similarities. We collect captions for motion pairs with middle similarity and employ ChatGPT~\cite{ouyang2022instructgpt,openai2023gpt4} to generate motion editing instructions that can transform one motion into another. Furthermore, we manually construct highly similar motion pairs for motion length editing tasks based on their respective lengths. By randomly combining these single-turn generation tasks, we can create a dialog format. The resulting tasks, along with diverse prompt instructions, are presented in ~\cref{tab:appendix:dataset}. We will release the pre-processed dataset.

\section{Motion Representations}
\label{sec:appendix:motionRepre}
We summarize two kinds of motion representations as follows.

\textbf{HumanML3D Format}~\cite{Guo_2022_CVPR_humanml3d} introduces a motion representation $x^{1:L}$ that draws inspiration from motion features in character control~\cite{starke2019neural, 2021-TOG-AMP, starke2022deepphase}. This representation, which contains redundant information, is well-suited for neural models, particularly variational autoencoders. Specifically, the $i$-th pose $x^i$ is defined by a tuple consisting of the root angular velocity $\dot{r}^a \in \mathbb{R}$ along the Y-axis, root linear velocities $(\dot{r}^x, \dot{r}^z \in \mathbb{R})$ on the XZ-plane, root height $r^y \in \mathbb{R}$, local joint positions $\mathbf{j}^p\in\mathbb{R}^{3N_j}$, velocities $\mathbf{j}^v\in\mathbb{R}^{3N_j}$, and rotations $\mathbf{j}^r\in\mathbb{R}^{6N_j}$ in root space. Additionally, it includes binary foot-ground contact features $\mathbf{c}^f \in \mathbb{R}^4$ obtained by thresholding the heel and toe joint velocities. Here, $N_j$ represents the number of joints, yielding the following representation:
\begin{align}
x^i = \{\dot{r}^a, \dot{r}^x, \dot{r}^z, r^y, \mathbf{j}^p, \mathbf{j}^v, \mathbf{j}^r, \mathbf{c}^f\}.
\end{align}

\textbf{SMPL-based Format}~\cite{SMPL2015} is a widely used parametric human model, SMPL~\cite{SMPL2015}, and its variants~\cite{MANO:SIGGRAPHASIA:2017, SMPLX2019}, which propose motion parameters $\theta$ and shape parameters $\beta$. The rotation vectors $\theta \in \mathbb{R}^{3\times23+3}$ represent the rotations of joints and the root, while $\beta$ represents the weights for linear blended shapes. This representation is commonly employed in markerless motion capture~\cite{he2021challencap,chen2021sportscap,VIBE_CVPR2020}. By including the global translation $r$, the representation is formulated as:

\begin{align}
 x^i = \{r, \theta, \beta\}.
\end{align}

\begin{table}[t]
\centering
\resizebox{\columnwidth}{!}{%
\begin{tabular}{@{}lll@{}}
\toprule
Task & Input & Output
\\ \midrule
\multirow{3}{*}{Text-to-Motion}& Show me a sequence of movements that illustrates [caption]. &\multirow{3}{*}{[motion] }
\\ & Demonstrate a motion that symbolizes the input: [caption]. &
\\ & I need a human motion that represents [caption].&
\\
\midrule \multirow{2}{*}{Text-to-Motion w/ length}& Please generate a motion that is around [frames] frames long for the caption: [caption]. &\multirow{2}{*}{[motion] }
\\ & Generate a motion that lasts for [seconds] seconds, and captures the essence of [caption]. &
\\ 
\midrule \multirow{2}{*}{Motion-Length-Editting }& Extend the duration of the motion provided. &\multirow{2}{*}{[motion] }
\\ & Reduce the duration of the motion without losing its main characteristics and precision. &
\\ \midrule \multirow{2}{*}{Length-to-Motion }& I want to see a motion that lasts for [frames] frames. &\multirow{2}{*}{[motion] }
\\ & Show me a motion that has a duration of [seconds] seconds. &
\\ \midrule \multirow{2}{*}{Radnom Motion }& Just show me a moving human. &\multirow{2}{*}{[motion] }
\\ & Produce motions that are not planned or choreographed.. &
\\ \midrule \multirow{2}{*}{Motion-to-Text}& Provide a description of the motion shown in [motion] using natural language. &\multirow{2}{*}{[caption] }
\\ & Provide a text-based explanation of what is happening in [motion]. & 
\\ \midrule \multirow{2}{*}{Motion-to-Text w/ length}& Generate a text summary for the [motion] that takes [frames] seconds to complete. &\multirow{2}{*}{[caption] }
\\ & Describe the movement exhibited in [motion] that is shown for a length of [seconds] seconds? &
\\ \midrule \multirow{2}{*}{Motion-to-Length}& How long does [motion]'s poses last in seconds?? & There are [frames] frames in the motion.
\\ & Calculate the second duration for [motion]'s body movements in seconds? & The motion lasts for [seconds] seconds.
\\ \midrule \multirow{2}{*}{Caption-to-Length}&HPredict the anticipated frame duration for the motion that corresponds to [caption]?& The duration is estimated to be around [frames] frames.
\\ & Guess the second count required for the motion represented by [caption]. & The motion has a length of [seconds] seconds.
\\ \midrule \multirow{2}{*}{Length-to-Caption}&Given the [frames] frames of the motion, what are some possible actions that could be taken?&\multirow{2}{*}{[caption] }
\\ & [seconds] is the number of motion seconds, generate the motion description: &
\\ \midrule \multirow{2}{*}{Random Caption}&Depict a motion as like you have seen it.&\multirow{2}{*}{[caption] }
\\ & Describe the motion of someone randomly.&
\\ \midrule \multirow{2}{*}{Motion-Reasoning }& 
Can you tell me what muscles are being used during this motion? & 
\begin{tabular}[c]{@{}l@{}}
This motion primarily targets the quadriceps, hamstrings, glutes,  \\and core muscles. It also engages the shoulders and upper back \\ muscles while raising the arms.
\end{tabular}
\\\cmidrule(lr){2-3}
& What could be the reason for the person not swinging their arms while walking? &
\begin{tabular}[c]{@{}l@{}}
There could be various reasons for this, such as the person carrying \\ something heavy or trying to maintain a certain posture while walking.
\end{tabular}
\\ \bottomrule
\end{tabular}%
}
\vspace{5pt}
\caption{A few examples of prompt templates used in our standardized motion conversation evaluation protocol.}
\label{tab:appendix:dataset}
\vspace{-10pt}
\end{table}

\section{Metric Definitions}
\label{appedix:metrics:details}
In the following section, we present additional details regarding the evaluation metrics.

\textbf{Linguistic Quality}.
To evaluate motion question-answer tasks, we employ linguistic metrics that assess the degree of alignment between the generated results and the ground-truth labels. These metrics include BLUE~\cite{papineni2002bleu}, Rouge~cite{lin2004rouge}, Cider~\cite{vedantam2015cider}, and BertScore~\cite{zhang2019bertscore}. For detailed information, please refer to the respective papers associated with each metric.

\textbf{Motion Quality}.
The Frechet Inception Distance (FID) serves for evaluating the distribution similarity between generated and real motions. It is calculated using a suitable feature extractor~\cite{guo2020action2motion,petrovich21actor,Guo_2022_CVPR_humanml3d} specific to each dataset. Additionally, we employ popular metrics in motion capture~\cite{VIBE_CVPR2020, chen2021sportscap, vonMarcard2018}, such as MPJPE and PAMPJPE~\cite{gower1975generalized}, to measure global and local errors in millimeters. To assess temporal quality, we utilize the Acceleration Error (ACCL). Furthermore, in line with previous motion prediction studies~\cite{yuan2020dlow, zhang2021we, ma2022multi}, we define the Average Displacement Error (ADE) as the average L2 distance between the ground truth and predicted motion for the entire sequence. The Final Displacement Error (FDE) is calculated as the L2 distance between the ground truth and predicted motion in the last frame.

\textbf{Motion Diversity}. 
Following previous studies~\cite{guo2020action2motion,chuan2022tm2t,mdm2022human}, we employ two metrics, Diversity (DIV) and MultiModality (MM), to evaluate the variability of motion across the entire dataset and the diversity of generated motion within each text input, respectively. To assess Diversity, the generated motions are randomly divided into two equal-sized subsets, and the Diversity metric is computed as the average distance between the motions in these subsets. For MultiModality evaluation, a set of text descriptions is randomly sampled from the available descriptions. Each text description is then replicated $m$ times for motion generation, and the MultiModality metric is defined as the average distance between the motions generated from the same text description.

\textbf{Condition Matching}.
HumanML3D~\cite{Guo_2022_CVPR_humanml3d} and TMR~\cite{petrovich2023tmr} provide motion/text feature extractors that generate geometrically coherent features for aligned text-motion pairs and vice versa. Within this feature space, we assess the motion-retrieval precision (R Precision) by combining the generated motion with 31 mismatched motions and calculating the top-1/2/3 matching accuracy between the text and motion. Additionally, we measure the Multi-modal Distance (MM Dist), which quantifies the distance between the generated motions and the corresponding text..

\textbf{Time Costs}. 
To assess the computational efficiency of our models, particularly the inference efficiency, we measure the average Frames Per Second (FPS) during motion generation. Specifically, we calculate the FPS on the test set of HumanML3D~\cite{Guo_2022_CVPR_humanml3d}, with a batch size of one, while excluding the time required for model and dataset loading.